\documentclass[conference,compsoc]{IEEEtran}

\usepackage[nocompress]{cite}

\usepackage{tikz}
\usepackage{amsmath}
\usepackage{array,booktabs,arydshln,color, caption}
\usepackage{amsmath,bm}
\usepackage{multirow}
\usepackage{caption}
\usepackage[ruled, lined, linesnumbered, commentsnumbered, longend,vlined]{algorithm2e}
\usepackage{algpseudocode} 
\usepackage{subcaption}
\usepackage{dblfloatfix}
\usepackage[hidelinks]{hyperref}
\usepackage{url}

\usepackage{graphicx,calc}
\usepackage{scalerel,amssymb}

\usepackage[symbol]{footmisc}

\usepackage{tikz,eso-pic}

\newlength\myheight
\newlength\mydepth
\settototalheight\myheight{Xygp}
\newcommand*\inlinegraphics[1]{%
  \settototalheight\myheight{Xygp}%
  \settodepth\mydepth{Xygp}%
  \raisebox{-\mydepth}{\includegraphics[height=\myheight]{#1}}%
}

\SetCommentSty{mycommfont}

\newcommand{\ProjectName}[1]{{\small\textsc{HateGuard}}} 
\newcommand{\ProjectNameTable}[1]{{\small\textsc{\textbf{HateGuard}}}} 

\newcommand{\meta}[1]{\textcolor{black}{#1}}
\newcommand{\crv}[1]{\textcolor{black}{#1}}

\newcommand{\NishNDSS}[1]{\textcolor{black}{#1}}

\newcommand{\placeholder}[1]{\textcolor{red}{XX}}

\begin{document}

\date{}




\title{Moderating New Waves of Online Hate \\with Chain-of-Thought Reasoning in Large Language Models}

\author{
\IEEEauthorblockN{Nishant Vishwamitra\IEEEauthorrefmark{1}\IEEEauthorrefmark{5},
Keyan Guo\IEEEauthorrefmark{2}\IEEEauthorrefmark{5}, 
Farhan Tajwar Romit\IEEEauthorrefmark{1}, 
Isabelle Ondracek\IEEEauthorrefmark{2},\\
Long Cheng\IEEEauthorrefmark{3},
Ziming Zhao\IEEEauthorrefmark{2},
Hongxin Hu\IEEEauthorrefmark{2}
}
\IEEEauthorblockA{
\IEEEauthorrefmark{1}University of Texas at San Antonio, 
\IEEEauthorrefmark{2}University at Buffalo,
\IEEEauthorrefmark{3}Clemson University\\
\{nishant.vishwamitra, farhantajwar.romit\}@utsa.edu, lcheng2@clemson.edu\\
\{keyanguo, ikondrac, zimingzh, hongxinh\}@buffalo.edu
}}

\maketitle
\begin{tikzpicture}[remember picture, overlay]
  \node[minimum width=4in] at ([yshift=-1cm]current page.north)  {To Appear in the 45th IEEE Symposium on Security and Privacy, May 20-23, 2024.};
\end{tikzpicture}

\footnotetext{\IEEEauthorrefmark{5}These authors contributed equally to this work.}
\renewcommand*{\thefootnote}{\arabic{footnote}}


\begin{abstract} 
Online hate is an escalating problem that negatively impacts the lives of Internet users, and is also subject to rapid changes due to evolving events, resulting in new waves of online hate that pose a critical threat. Detecting and mitigating these new waves present two key challenges: it demands reasoning-based complex decision-making to determine the presence of hateful content, and the limited availability of training samples hinders updating the detection model. To address this critical issue, we present a novel framework called \ProjectName{} for effectively moderating new waves of online hate. 
\ProjectName{} employs a reasoning-based approach that leverages the recently introduced chain-of-thought (CoT) prompting technique, harnessing the capabilities of large language models (LLMs). \crv{\ProjectName{} further achieves prompt-based zero-shot detection by automatically generating and updating detection prompts with new derogatory terms and targets in new wave samples to effectively address new waves of online hate.} To demonstrate the effectiveness of our approach, we compile a new dataset consisting of tweets related to three recently witnessed new waves: the 2022 Russian invasion of Ukraine, the 2021 insurrection of the US Capitol, and the COVID-19 pandemic. Our studies reveal crucial longitudinal patterns in these new waves concerning the evolution of events and the pressing need for techniques to rapidly update existing moderation tools to counteract them. Comparative evaluations against state-of-the-art approaches illustrate the superiority of our framework, showcasing a substantial 10.59\% to 88\% improvement in detecting the three new waves of online hate. Our work highlights the severe threat posed by the emergence of new waves of online hate and represents a paradigm shift in addressing this \crv{threat} practically. \looseness=-1

\end{abstract}
\vspace{0.2cm}
\noindent \textbf{Disclaimer}. This manuscript contains harmful content, including hate speech, which has the potential to be offensive and may disturb readers.

\thispagestyle{empty}
\section{Introduction}


We live in a world with rapidly evolving events. These rapidly evolving events consequently affect the global digital landscape~\cite{valdez2020social}, especially Internet platforms that enable online discourse, such as Online Social Networks~(OSNs). As a result, emotions of anger and anxiety, and rhetoric from these events also spill over into our global digital landscape. For example, recent polarizing events, such as the 2022 Russian invasion of Ukraine~\cite{russia_ukraine_hate}, the 2021 insurrection of the US Capitol~\cite{capitol_hate}, and the COVID-19 pandemic~\cite{he2021racism}, rapidly transformed the online discourse in our cyberspaces~\cite{tsunami_of_hate}. As a consequence, the context of online hate has also rapidly changed, leading to the emergence of \textit{new waves} of online hate. For example, during the 2021 insurrection of the US Capitol, a wave of hateful content against vulnerable groups, such as LGBTQ and minorities, was witnessed~\cite{prabhu2021capitol,capitol_hate}. During the COVID-19 pandemic, a rapid rise in online hate against Asian-Americans~\cite{he2021racism,chiang2020anti}, mask~\cite{choi2021mask,wang2020public}, and vaccine mandates~\cite{wardle2021too,broniatowski2021first} was reported on several OSNs. More recently, during the 2022 Russian invasion of Ukraine, we saw yet newer waves of online hate against citizens of the nations involved in the conflict~\cite{wypych2022psychological,shevtsov2022twitter}. New waves of online hate are a crucial issue that demands immediate attention not only for the present but also for the future well-being of our digital spaces. As new waves are likely to arise in the future, it becomes imperative to address this problem proactively.\looseness=-1


\looseness=-1

Currently, various existing tools, such as Perspective API~\cite{PerspectiveAPI}, Azure Text Moderation~\cite{azure_text_moderation}, and IBM Toxic Comment Classifier~\cite{ibm_toxic_comment_classifier}, utilize artificial intelligence and machine learning~(AI/ML) models for moderating violations of online hate 
policies~\cite{fb_hate_speech_def,twitter_hate_speech_policy}. 
However, there is a concern regarding the effectiveness of these tools in preventing violations caused by new waves of online hate. For example, the recent Anti-Asian hate~\cite{he2021racism,chiang2020anti}, mask-related hate~\cite{choi2021mask,wang2020public}, and vaccine-related hate~\cite{wardle2021too,broniatowski2021first} witnessed during the COVID-19 pandemic could not be sufficiently detected by these tools, and online hate against minority communities and other vulnerable groups continued to spread unabated during this period. 
%
%

A major limitation of these existing tools, which hinders their effectiveness against new waves of online hate, is their reliance on traditional AI/ML models. This poses two key challenges.
\textit{First}, the detection of new waves of online hate poses a complex decision-making challenge, significantly different from the traditional classification tasks typically addressed by AI/ML models. Online hate is inherently ``highly subjective, ambiguous, and context-dependent, making it difficult for both humans and computers to detect''~\cite{davidson2022machine}, and the emergence of new waves exacerbates these difficulties. For instance, during the COVID-19 pandemic, a new wave of online hate targeted emerging political identities like ``antimaskers'', employed novel disparaging terms like ``maskhole'', and utilized different stereotypes to target specific communities.
Determining whether such content is hateful or not demands intricate decision-making that necessitates reasoning. This process involves exploring multiple possibilities, including carefully discerning between expressions of hateful speech, mere criticism, and ironic statements~\cite{gelber2002speaking,howard2019free}.
\textit{Second}, due to the abrupt occurrence of new waves, only a limited number of samples are accessible for model updates. Therefore, tools designed to detect and moderate this issue should possess the capability for rapid deployment and adaptation, utilizing either minimal or no samples of new waves. Nevertheless, existing tools face challenges in promptly adjusting to the sudden surge of a new wave, as they lack a sufficient number of training samples. Additionally, the training paradigm employed by these tools necessitates the collection of a large dataset, followed by manual labeling through human annotators, a process that typically takes months and is not practically feasible for the timely discovery and moderation of new waves of online hate~\cite{fb_evolving_hate}. 



In this work, we embark on addressing the practical challenges presented by new waves of online hate. To begin, we analyze several recently emerged new waves, specifically, the 2022 Russian invasion of Ukraine, the 2021 US Capitol insurrection, and the COVID-19 pandemic. To support our research, we gather a novel dataset,
containing 31,549 tweets related to these three categories of new waves. We present two systematic studies examining the nature of these new waves and the necessity for novel methods to update existing moderation tools. The first study focuses on tracking the usage of hateful hashtags associated with the three new waves in our dataset, revealing significant longitudinal patterns that can be leveraged for rapid detection. Subsequently, we explore the effectiveness of techniques employed to update existing online hate moderation tools against multiple new waves of online hate. Our findings reveal that these techniques fail to address the challenges presented by the emergence of new waves.

Based on these findings, we design \ProjectName{}\footnote{Our code and datasets are available at \url{https://github.com/CactiLab/HateGuard}.}, a novel framework for the discovery and moderation of new waves of online hate.
 \ProjectName{} introduces a reasoning-based approach, capitalizing on the recent innovation of chain-of-thought (CoT)  prompting~\cite{wei2022chain}, enabling large language models (LLMs)  to undertake the complex decision-making task of identifying whether new content is hateful or not. \crv{Additionally, \ProjectName{}  employs an automatic strategy to generate and update prompts for zero-shot classification by only updating the newly identified hate targets and derogatory terms in the prompts rather than the model.} Our approach tackles the challenge of detecting new waves of hate by carefully crafting chains of automatic reasoning through LLMs, probing, and exploring various possibilities within the content. This methodology proves more suited to the intricate decision-making requirements of detecting new waves compared to traditional binary classification approaches. \crv{Moreover, our zero-shot approach facilitates updates solely to the prompts while leaving the model untrained.}
 \looseness=-1

The key contributions of this paper are as follows:
\begin{itemize}
  \item \textbf{New dataset of new waves of online hate.} To study and understand the nature of the new waves of online hate, and to demonstrate the effectiveness of our approach, we collect a new dataset of 31,549 tweets about three recent new waves of online hate: the 2022 Russian invasion of Ukraine, the 2021 US Capitol insurrection, and the COVID-19 pandemic.  

  \item \textbf{\NishNDSS{New understanding about new waves of online hate.}} We report two systematic studies on the nature of new waves of online hate and the need for techniques to rapidly update existing online hate moderation tools against the new waves. Our studies shed light on longitudinal patterns regarding the sharp rise, peak, and dissipation of these new waves with evolving events that can be leveraged to rapidly detect such new waves, and highlight the need for methods to quickly update existing moderation tools.

  \item \textbf{New framework for the moderation of new waves of online hate.} 
  We design a novel framework called \ProjectName{} for effectively moderating new waves of online hate.  \crv{\ProjectName{} incorporates a CoT reasoning approach, empowering LLMs with reasoning capabilities to determine whether new content exhibits hateful characteristics, and an automatic prompt generation and update strategy for zero-shot classification, which streamlines the update process by solely focusing on updating the prompts rather than the model.} Our framework takes a first step towards practically moderating new waves of online hate, by harnessing the potency of LLMs.
  \item \textbf{Multi-faceted and extensive evaluation of \ProjectName{}.}  
  We showcase \ProjectName{}'s capability to enhance flagging of different types of new waves, achieving an impressive improvement ranging from 6.52\% to 71.93\% compared to baselines in the last quarter of these new waves. Additionally, we compare our framework against state-of-the-art models and demonstrate its superiority, achieving 10.59\% to 88\% higher accuracy than these models.
  We also apply our framework in a real-world scenario, where it effectively identifies and flags \textit{all} the hateful samples within a dataset of in-the-wild samples.\looseness=-1
\end{itemize}


\section{Background and Related Work}

In recent times, online hate has emerged as a critical threat~\cite{thomas2021sok} that has been the focus of both governments~\cite{doj_hate_crime_laws} and institutions~\cite{twitter_hate_speech_policy,fb_hate_speech_def}. It has been reported that in 2017, 41\% of Americans reported personally experiencing varying degrees of harassment online~\cite{duggan2017online}, and  40\% of users reported similar experiences globally~\cite{microsoft_civility}. 
Furthermore, new vectors of online hate~\cite{qu2022evolution} evolve fast, thus evading existing detection systems. 
Understanding the need to address this growing threat, concerned persons from both academia~\cite{davidson2017automated} and industry~\cite{facebook_hate_speech_AI,twitter_hate_speech_AI} have made efforts to defend against this threat. Initially, methods that involve human moderators have been proposed~\cite{fb_community_standards_enforcement_report,healthier_twitter}, although the practicality of scaling these methods is questionable, and these methods are also not ethical~\cite{content_moderators}. AI/ML has since emerged as a critical technology that is being explored to practically address this threat. Recent studies have employed AI/ML techniques to develop classifiers capable of moderating online hate speech~\cite{dinakar2011modeling,warner2012detecting}. However, these approaches cannot be used for the discovery and moderation of new waves of online hate.\looseness=-1

\NishNDSS{The new waves of online hate can be considered as a specific case of concept drift~\cite{lu2018learning} in the domain of online hate. Concept drift is defined as the ``changes in the hidden context that can induce more or less radical changes in the target concept''~\cite{widmer1996learning}. The concept drift problem is critical in AI/ML since changes in the target's statistical properties can render a model less effective or even useless~\cite{tsymbal2004problem}. Although the presence of the concept drift problem has been discussed in the context of online hate~\cite{omar2022making}, methods that specifically address it in the online hate domain have not been yet developed. Our framework offers a potential solution for addressing the issue of concept drift in online hate.}


A critical issue with online hate detection is that it is a complex decision-making problem~\cite{davidson2022machine,gelber2002speaking,howard2019free}. While ML algorithms perform remarkably well on classification tasks, they are not well suited for decision-making tasks that demand reasoning~\cite{macavaney2019hate}. 
Recent research~\cite{davidson2022machine} indicates that online hate is highly contextual and poses a significant challenge for moderation, even for humans. Facebook has acknowledged that human moderators are essential and always have the final say when determining if flagged posts should be removed for hate content~\cite{facebookalwayshuman}.
New waves of online hate compound this issue. Since there are significant differences between the semantics of new waves, such as the use of new derogatory terms and new targets of hate, hate speech detection models' performance further deteriorates when faced with new waves of hate~\cite{he2021racism}. As a result, there is a need for reasoning-based approaches to identify new hateful content, involving decision-making to determine whether the content is hateful or not. \crv{Recently, in-context learning~(ICL)~\cite{nye2021show,wei2022chain} has emerged as a method to learn a new task from a small set of examples presented within the context (the prompt) at inference time. ICL enables pre-trained LLMs to address new tasks without the need for fine-tuning. Especially, The adoption of chain-of-thought~(CoT) prompting in LLMs~\cite{wei2022chain}, as an ICL technique, has given them reasoning capabilities, opening up a new era in decision-making AI.} While a recent study has applied such a technique for explaining AI decisions~\cite{huang2022chain}, the specific challenges associated with using the CoT approach for moderating online hate remain unexplored. \looseness=-1

\begin{figure}[t]
\vspace{-0.1cm}
\centering
\resizebox{1\columnwidth}{!}{\includegraphics [width=1\linewidth]{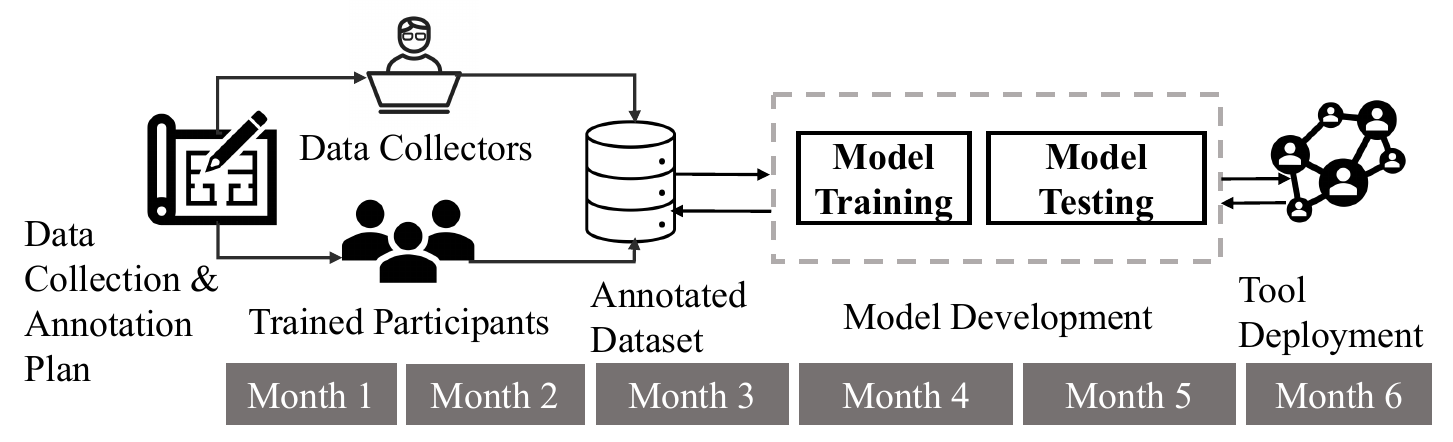}}
\vspace{-0.3cm}
\caption
{Conventional AI training approach.}
\label{fig:existing_tools_method}
\vspace{-0.2cm}
\end{figure}

Furthermore, existing approaches against new waves follow a conventional AI training approach that is illustrated in Figure~\ref{fig:existing_tools_method}, which is not practical for the real-world discovery of new waves, and cannot support quick deployment, because this process is quite time-consuming~\cite{fb_evolving_hate}. In the conventional approach, first, a data collection and annotation plan is designed that describes what kind of posts should be deemed as hateful~\cite{davidson2017automated,dinakar2011modeling}. Then, the data collection is done wherein such posts are collected via social media APIs~\cite{dinakar2011modeling,vishwamitra2020analyzing}. Next, humans are trained to perform annotation tasks on the collected dataset, and after this process, we get an annotated dataset that can be used to train AI models. In the next step, the AI model is trained on the annotated dataset and the performance is evaluated on a test dataset. If the performance is satisfactory, it is deployed on Internet platforms for discovery and moderation. However, a major issue with this approach is that it takes \textit{months} to complete this process~\cite{fb_evolving_hate},  which makes it impractical to address the problem of new waves of online hate that need rapid updating of the model. \looseness=-1

Few-shot and zero-shot learning (FSL and ZSL)~\cite{wang2019survey} have recently emerged as a way of developing AI models using a few or no data samples. However, these techniques have been predominantly applied for images, such as adding a new category of an object to an existing dataset to enhance a model's detection capability. Recently, the use of FSL has been explored in text-based applications~\cite{yan2018few}, and some emergent studies have explored FSL for detecting hate speech in less common languages~\cite{stappen2020cross,mozafari2022cross} and task decomposition~\cite{alkhamissi2022token}. For example, ~\cite{stappen2020cross,mozafari2022cross} discuss approaches to detect hate speech in rare languages using transformer-based models in a few-shot setting (\textit{i.e.}, simulating a small number of samples). However, they don't provide specific approaches for FSL to address online hate. 
Recent studies have shown that LLMs can even outperform humans in zero-shot tasks~\cite{bang2023multitask}, and in this work, we explore the zero-shot capability of LLMs in moderating new waves of online hate. \looseness=-1




\section{Threat Model}


In this work, we address the behavior of adversaries who spread hate online, especially those who target individuals or groups based on their identity, often in reaction to evolving events. Both the adversary and the target are considered as online users. The adversary can create hateful posts using various constructs, including words and hashtags relevant to the evolving events. 
We focus solely on textual media and do not consider other means of disseminating such posts, such as images, upvotes, or likes.
Our study addresses posts targeted specifically against a particular user, as well as posts intended for a wider audience, such as public posts. We make no assumptions about the adversary possessing any special capabilities or employing adversarial techniques to deceive content moderation tools.

\section{Examining New Waves of Online Hate}
\label{sec:examining}
In this section, we present studies on the nature of the new waves of online hate considering three recent new waves. In the first study, we investigated how new waves of online hate emerged with the changes in the global digital landscape. \NishNDSS{In the second study, we examined the need to quickly update existing tools used for online hate discovery and moderation by measuring their moderation capability on new waves of online hate. Our main objectives in conducting these two studies are to find out if there are patterns in the nature of the new waves that could be utilized for their detection and motivate the need for new strategies to quickly enable existing approaches to handle new waves.}


\subsection{Data Collection and Annotation}
\label{subsec:datacollection}

Our data collection and annotation tasks were approved by our Institution's IRB. We carried out two dataset tasks, collection and annotation. To collect tweets related to the three recent new waves, we compiled a \crv{seed} set of hashtags that were prevalent during the time these waves~\cite{trumpchinesevirus.2020, covidiot.2020,covidiottrend.2020} were active, which consisted of diverse hashtags such as \textit{\#ChinaVirus}, \textit{\#WuhanFlu}, \textit{\#WearAMask}, \textit{\#boomerremover}, \textit{\#COVID19Vaccine}, \textit{\#MAGAMorons}, and \textit{\#F**Putin}. \crv{We expanded this set by adding new hashtags from the collected tweets until no new ones were found, ensuring a representative sample.} The full list of hashtags has been provided in Appendix~\ref{subsec:hashtags_list}.

\noindent \textbf{Collection of Tweets.} 
We used the official X (previously Twitter) Streaming API\footnote{https://developer.twitter.com/en/docs/twitter-api} to collect tweets during the period from December 1, 2019, to December 31, 2022, based on the hashtags. Particularly, we collected COVID-19-related tweets from December 1, 2019, to December 31, 2020, US Capitol insurrection-related tweets from November 1, 2020, to December 31, 2021, and tweets regarding the Russian invasion of Ukraine from November 1, 2021, to December 31, 2022. In total, we obtained 507 million tweets published by 38 million users. We removed tweets with only hashtags, mentions, or links in the text part, and removed non-English tweets and retweets. We were left with 31,549 tweets. Additional samples from our dataset can be viewed in Appendix~\ref{app:samples}.\looseness=-1

\noindent \textbf{Annotation of Tweets.} Two authors of this work developed a code book for labeling the samples in our collection as hate speech or not and verified it after three rounds of annotations and resolving conflicts on random samples of the dataset. \crv{The two authors independently annotated 300 random samples from our dataset in each round, followed by agreement computations and conflict resolution discussions. This process was repeated with different random samples. By the third round, the two authors achieved 100\% agreement. } To develop the code book, we focused on identity-based hate and hate against individuals, defined as ``Hatred, hostility, or violence towards member(s) of a race, ethnicity, nation, religion, gender, gender identity, sexual orientation or any other designated sector of society''~\cite{macavaney2019hate}. To ensure the accuracy of the annotations, we meticulously cleaned each tweet by removing URLs, mentions, and non-English characters. Additionally, we removed stacked hashtags as well as those at the beginning and end of a tweet, unless they are linked to action words or determiners like ``a'',``an'', or ``the''~\cite{bao2014role}.
In our code book~(detailed fully in Appendix~\ref{app:mturkannotations}, Table~\ref{tab:codebook}), our analysis began with identifying if an individual or group identity is mentioned in the text. We then assessed for any derogatory or disparaging language, followed by determining if such language was directed at the mentioned individual or identity. We used Amazon Mechanical Turk~(AMT) to label a random sample of 4,000 tweets using online participants and directed workers to label a text as hate if such words were directed at the identities or individuals mentioned. We labeled a subset of our dataset since we found that our task is quite intensive since it needs a lot of human reasoning and time, which was not practical to extend to the entire dataset. \meta{Furthermore, we sampled this subset with a temporal distribution, \textit{i.e.}, we proportionally sampled an equivalent number of tweets from each quarter. Overall, we sampled 928 tweets from Q1, 893 tweets from Q2, 1,148 tweets from Q3 and 1,031 tweets from Q4.} To maximize reliable annotation, we only recruited participants with an approval rating of 90\% or higher and 1,000 approved HITs to participate in our annotation task. \crv{The AMT workers on average labeled 32 tweets.} Since one of the new waves, the US Capitol insurrection, is linked to US politics, we have chosen to limit the geographical scope of our study to the US and Canada. After the annotation task was completed, the two expert annotators and developers of the code book who are well-trained in using it to label the samples verified the labels and corrected any coding errors made by the workers. The experts achieved a Fleiss Kappa agreement of 0.84, which indicates a near-perfect agreement. Table~\ref{tab:dataset} shows the results of our tweets collection task. Of the 4,000 tweets labeled, 1,647 were labeled as hate and 2,353 as non-hate. Additionally, we were left with a large dataset of the rest of the 27,549 tweets to support large-scale analysis.\looseness=-1


\begin{table}[t]
\centering
\resizebox{\columnwidth}{!}{\begin{tabular}{ccc}
\toprule
\textbf{New Wave Type} & \textbf{\begin{tabular}[c]{@{}c@{}}Number of \\ hateful tweets\end{tabular}} & \textbf{\begin{tabular}[c]{@{}c@{}}Number of non \\ hateful tweets\end{tabular}} \\
\midrule
COVID-19 tweets & 1,096 & 1,600 \\ 
US Capitol Insurrection tweets & 314 & 390 \\ 
Russian Invasion of Ukraine tweets & 237 & 363\\
\midrule
Total tweets & 1,647 & 2,353\\
\bottomrule
\end{tabular}}
\caption{Annotated new wave dataset with 4,000 tweets.}
 \vspace{-4mm}
\label{tab:dataset}
\end{table}

\subsection{Nature of New Waves of Online Hate}
\label{subsec:nateuofnewwaves}


\begin{figure*}[t!]
    \centering
    \includegraphics[width=\textwidth]{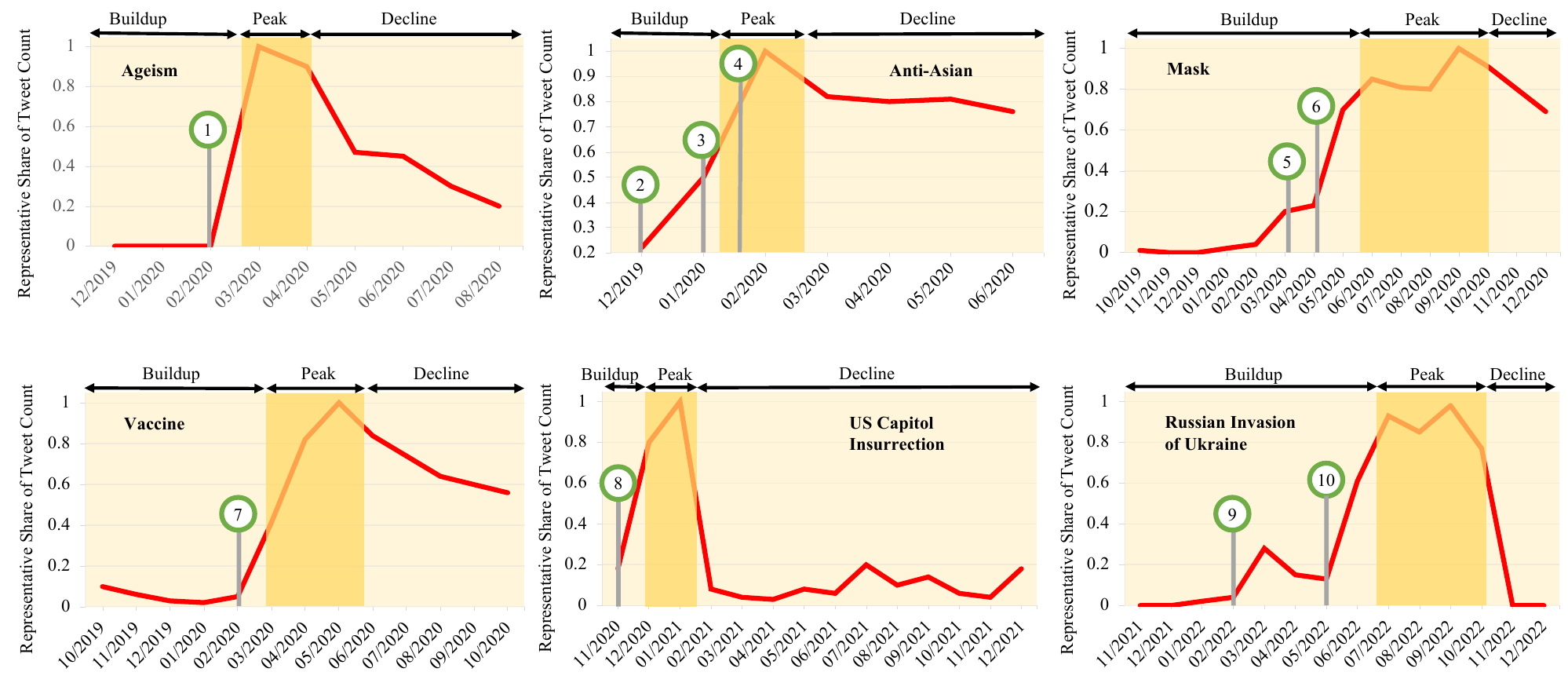}
    
    \vspace{-3mm}
    \caption{As current events evolve, new waves of online hate occur in the global digital landscape. }
    \label{fig:hate_context_evolution}
    \vspace{-5mm}
\end{figure*}


To understand the nature of online hate, we investigated the text-based tweets by studying the temporal usage pattern of the extremely hateful X hashtags (such as \#WuhanFlu, \#boomerremover, \#F**Putin, \#MAGAMorons, etc.) about the three new waves in our dataset. 
Furthermore, during the COVID-19 pandemic, several distinct new waves of online hate emerged. Our study specifically examined four categories prevalent in social media during this time~\cite{he2021racism,taylor2021negative}: Anti-Asian, Ageism, Mask, and Vaccine. We specifically focused on a ``wave'' of these new categories of hate, \textit{i.e.}, a sharp, sudden, or unprecedented increase of these tweets, and what the antecedents of this increase could be. Our observations are depicted in Figure~\ref{fig:hate_context_evolution} and Table~\ref{tab:new_wave_current_events}. \crv{We utilized the pruned exact linear time (PELT) algorithm~\cite{killick2012optimal} to find the change points in Figure~\ref{fig:hate_context_evolution}. Two authors then correlated these points with real-world events in Table~\ref{tab:new_wave_current_events}.} Figure~\ref{fig:hate_context_evolution} illustrates the temporal usage of hateful hashtags in each category, along with the corresponding dates of the events, and Table~\ref{tab:new_wave_current_events} displays the current events associated with each category. We observed that for each of the categories, there are three stages, \textit{buildup}, \textit{peak}, and \textit{decline},  in the temporal usage~(Figure~\ref{fig:hate_context_evolution}), which are closely related to the current events. 

\noindent \textbf{Buildup.} This is the stage when certain current events related to a category are building up emotions of hate, anger, and anxiety in social media. We observed that the posts in this stage were responsible for building an outbreak of hate at a later stage. In particular, awareness about a certain event also plays a major part in the buildup stage. For example, we observed that negative emotions in the Asian community were being built up due to certain events such as the US CDC screening people traveling from China~\cite{cdc_museum}, and the WHO issuing a Global Health Emergency~\cite{cdc_museum}, which added to the stress and strain due to imposed lock-downs. Events that built up negative emotions against the older individual were observed in this stage, such as reports across the world about the pandemic disproportionately affecting older individuals and subsequent tweets that used particularly offensive terms such as ``BoomerRemover'' for COVID-19~\cite{lichtenstein2021coffin}. 
The imposition of mask mandates across various institutions and public spaces~\cite{mask_mandates_2020}, coupled with the CDC's recommendations for mask-wearing, contributed to an increase in social media discussions and opinions about mask usage during this phase. In the case of vaccine-related hateful hashtags,  certain events, such as vaccine companies starting human trials of vaccines led to a buildup of emotions regarding the use of vaccines~\cite{vaccine_trials_start_2020}. We observed the buildup of such emotions in the case of US Capitol insurrection-related hate, wherein election day played a polarizing role~\cite{electionpolarizing}. Furthermore, events such as the day Russia invaded Ukraine and the siege of Mariupol~\cite{winter2022russian} (an event of significance during the invasion) witnessed a buildup of strong emotions.
The online activities in the \textit{buildup} stage are crucial since they are precursors for online hate. Suitable counter-actions in this stage are thus necessary to prevent an outbreak of online hate.\looseness=-1

\begin{table}[!b]
\centering
\vspace{-2mm}
\resizebox{\columnwidth}{!}{%
\begin{tabular}{ll}
\toprule
1 & COVID-19 referred to as ``BoomerRemover'' first time on Twitter \\ 
2 & WHO informs of cases of unexplained pneumonia in Wuhan, China\\ 
3 & US CDC starts screening people from China          \\ 
4 & WHO issues Global Health Emergency                                  \\ 
5 & Mask mandates put in place across various institutions              \\ 
6 & CDC recommends wearing masks                                        \\ 
7 & Companies start first vaccine trials                                \\
8 & US Elections                                \\
9 & Russia invades Ukraine                                \\
10 & Russia seizes Mariupol                                \\
\bottomrule
\end{tabular}%
}
\caption{Events that engendered new waves of online hate.}
\label{tab:new_wave_current_events}
\end{table}

\noindent \textbf{Peak.} In the following stage, an outbreak of usage of hateful hashtags was observed after the negative emotions built up in the previous stage led to a peak of the new waves of online hate. This stage depicted an uncontrolled and overwhelming usage of hateful hashtags and an apparent failure of OSNs in countering hateful activity. For example, a peak of Anti-Asian hate was observed between February 2020 and March 2020. In the case of Ageism, the peak was noted between March 2020 and April 2020. In the case of mask-related online hate, the peak was observed between June 2020 and October 2020. The peak for a vaccine-related wave of online hate was observed in May 2020. Lastly, the peak of US Capitol insurrection-related hate was observed in January 2021, and the peak of Russian invasion-related online hate was observed in September 2022. Pragmatic steps, especially in the preceding stage must be taken to avoid the \textit{peak} stage of a new wave of online hate. 

\noindent \textbf{Decline.} During this stage, there was a noticeable decrease in the use of hateful hashtags related to new waves of online hate on Twitter following their \textit{peak} stage. This decline could occur due to OSNs becoming aware of the new wave of online hate and taking measures, such as content moderation using human moderators~\cite{content_moderators}. But by the time this stage is reached, a large number of posts had already been shared on X in all the three types of new waves considered, including the four categories of COVID-19-related hate. Besides, in some categories, such as Anti-Asian, we observed that the \textit{decline} stage was much more gradual and prolonged than other categories (\textit{e.g.}, Ageism, Vaccine, and Russian invasion), showing sustained publishing of tweets that used hateful hashtags despite content moderation efforts.\looseness=-1



\NishNDSS{\textbf{Finding.} Our study reveals that real-world events can trigger new waves of online hate, leading to swift changes in hate speech dynamics. Our experiment has identified significant longitudinal patterns that can help address new waves of online hate.}  These new waves typically involve a \textit{buildup} of negative emotions resulting from evolving events, followed by a \textit{peak} stage where the outburst of the new wave is encountered, and then a \textit{decline} stage where the new wave reduces. We are developing a framework that can be updated based on only a few samples during the \textit{buildup} stage of a new wave, which then counters the \textit{peak} to significantly reduce its harmful effects. By doing so, our approach offers practical moderation of new waves of online hate.

\subsection{Using Existing Tools Against New Waves of Online Hate}

Following our previous study, we wanted to investigate the need for new methods to extend the capabilities of the existing moderation tools to new waves of online hate. Our objective is not to point out that these tools are not effective against new waves, but to motivate the need to quickly update these tools. Specifically, we wanted to motivate the need for methods that can address the issue of rapid concept drifts in online hate, and study the limitations of existing methods, such as fine-tuning, which is a popular method to update such tools. Although these models are proprietary black-box, it is quite likely that they are trained with fine-tuning-based strategies. For example, Perspective API uses a multilingual BERT, which primarily uses fine-tuning that is ``frequently retrained to make improvements and keep them up-to-date''~\cite{jigsawuptodate}. Perspective API~\cite{PerspectiveAPI} needs a dataset of at least 20,000 samples of a particular label to be considered enough to re-train a model~\cite{perspectiveapicontributefeedback}. Similarly, IBM Toxic Comment Classifier~\cite{ibm_toxic_comment_classifier} is based on fine-tuning a BERT-base uncased~\cite{ibmtoxiccommnetclassifierfinetuning} model, and likely uses the same strategy of fine-tuning with a comparatively large dataset to update their model.\looseness=-1

We measured several state-of-the-art tools (\textit{i.e.}, Clarifai Text Moderation~\cite{clarifai}, Perspective API, Azure Text Moderation~\cite{Azure}, and IBM Toxic Comment Classifier) against the hateful tweets in our dataset. Our objective in this measurement experiment was to study the capability of these existing systems on the new waves of online hate only, and we do not propose that these systems and models are not effective against hate in general, since they have been known to be effective against traditional hate~\cite{vidgen2020recalibrating}. We depict the results of this measurement experiment in terms of precision, recall, and F1-score in Table~\ref{tab:detection_capability_existing_systems}. We found that the existing tools are severely limited in discovering new waves of online hate observed from the low F1 scores reported by these systems. The highest F1 score was found to be just 0.38 (Perspective API), which is not sufficient for practical use.

\begin{table}[bt]
\centering
\resizebox{\columnwidth}{!}{%
\begin{tabular}{p{3.4cm}cccc}

\toprule
\textbf{Detection Tools}  
& \textbf{Precision} & \textbf{Recall} & \textbf{F1-score} \\ 

\midrule

Clarifai Text Moderation~\cite{clarifai}                    & \multirow{1}{*}{0.69}              & \multirow{1}{*}{0.16}           & \multirow{1}{*}{0.27}             \\
 
Perspective API~\cite{PerspectiveAPI}                                                         & 0.49               & 0.31            & 0.38              \\

Azure Text Moderation~\cite{azure_text_moderation}                                                  & \multirow{1}{*}{0.54}               & \multirow{1}{*}{0.21}            & \multirow{1}{*}{0.31}              \\

IBM Toxic Comment Classifier~\cite{ibm_toxic_comment_classifier}                                            & \multirow{2}{*}{0.69}              & \multirow{2}{*}{0.15}            & \multirow{2}{*}{0.25}              \\

\bottomrule
\end{tabular}%
}
\caption{Use of existing systems in detecting new waves of online hate. 
}
\label{tab:detection_capability_existing_systems}
\vspace{-4mm}
\end{table}

\textbf{Finding.} It is evident that an existing detection tool might not perform well when new waves of online hate are presented to them. However, they can be augmented by different means such as zero-shot (or few-shot) learning to adapt to such rapid changes in the concept. We acknowledge that other factors, such as tool owners not periodically updating their models, could also limit the tools' effectiveness against new waves. However, we focus solely on the limitations of existing tools due to the rapidly evolving nature of online hate. This limitation indicates that new methods for the discovery and moderation of new waves of online hate must be developed.

\section{\ProjectName{} Design}
\label{sec:approach}



\subsection{Design Intuition}

\begin{figure*}[t]
\vspace{-0.1cm}
\centering
\includegraphics [width=.95\linewidth]{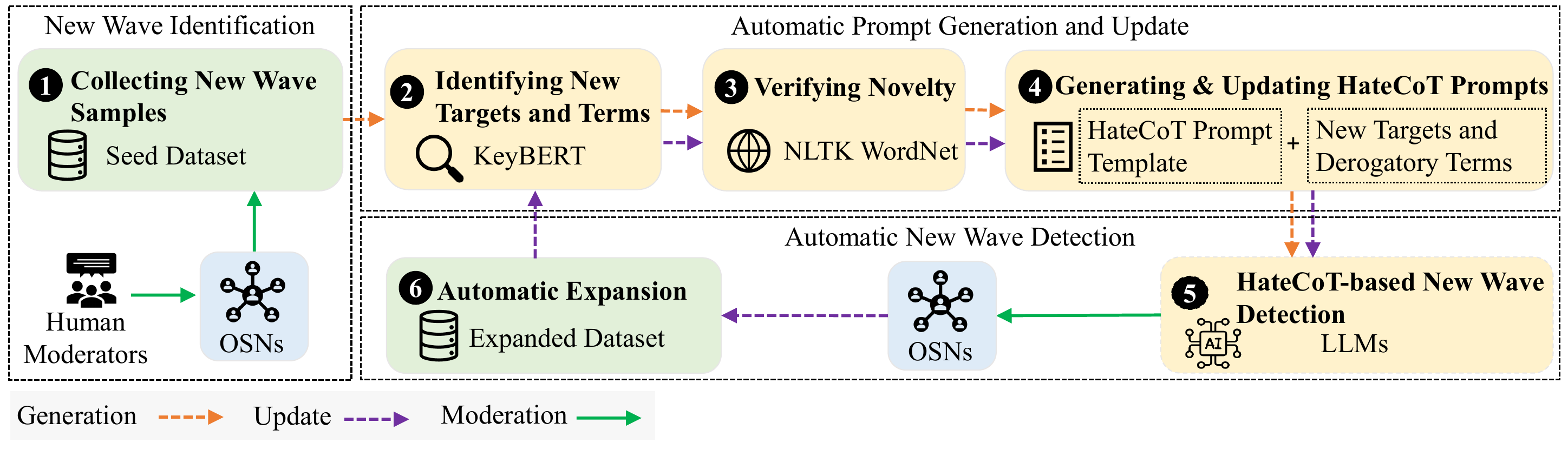}
\vspace{-0.2cm}
\caption
{Overview of \ProjectName{}.}
\vspace{-0.4cm}
\label{fig:overview}
\end{figure*}

Before delving into our approach to addressing new waves of online hate, we provide a discussion about the intuitions behind the design of our approach.

\noindent \textbf{Reasoning-based decision-making for detection.} Detection of hateful content is not a simple classification task, it is a complex decision-making task that involves reasoning~\cite{davidson2022machine}. A major reason for the complexity is due to its highly contextual nature. For instance, the decision whether or not new content is hateful is based on multiple factors and the interaction between these multiple factors. As an example, in identity-based hate~\cite{PerspectiveAPI}, the mere mention of an identity is not sufficient for it to be decided as hateful, there needs to be an element of an attack involving derogatory words towards the mentioned identity. Furthermore, the identity could in reality be an entity, such as the US government or the United Nations, and derogatory words being used to express mere criticism, in which case it is not hateful. 
The task's complexity is heightened by scenarios where derogatory words are used without targeting a specific identity, resulting in posts that are offensive but not necessarily hateful~\cite{davidson2017automated}. This complexity is so intricate that even individuals from diverse backgrounds often find it challenging to discern whether a text is truly hateful~\cite{hatespeechishardevenforhumans}. Moreover, the emergence of new waves adds another layer of complexity, introducing fresh contexts, such as novel derogatory terms and new targets of hate, which further muddle the process of determining hatefulness.\looseness=-1

The determination of whether new content is hateful or not is a complex and contextual decision-making process that is based on reasoning, and it is not a simple classification task. However, current ML-based techniques are based on the traditional paradigm based on training a model on a large hate speech dataset and making binary predictions. We argue that this paradigm is predominantly based on word associations in the training datasets, does not sufficiently consider context, does not consider hateful factors or the associations between them, and most importantly lacks reasoning-based decision-making.

The major challenge in performing this decision-making is how to practically control it on a large platform, such as social media. One option is to use human moderators. But using human moderators to achieve this goal is not suitable since humans' fatigue in such tasks~\cite{mao2013stop} and the extremely concerning ethical issues that human moderators face, such as post-traumatic stress disorders (PTSD) after viewing such content~\cite{steiger2021psychological}. Thus, the other option involving ML models is more suitable. However, current ML models are limited to classification tasks and are not suitable for complex reasoning-based decision-making. The recent invention of large language models (LLMs) has significantly revolutionized the landscape of NLP tasks~\cite{min2021recent}. Since these models are trained on massive amounts of data with a reinforcement paradigm, they can sufficiently capture the contextual information needed for NLP tasks and can be prompted to perform various tasks in a few or zero-shot manner. However, to develop LLMs to do reasoning-based tasks such as determination of hate, they can be prompted in several intermediate steps to arrive at the final decision based on the intermediate outputs, a recently introduced process known as \emph{chain-of-thought} (CoT) prompting~\cite{wei2022chain}. LLMs based on this prompting style have been shown to perform better on reasoning tasks, such as arithmetic problem solving~\cite{yao2023tree,fu2023chain}. However, a CoT prompting process for online hate detection is yet unexplored. These intermediate prompts need to be thoughtfully designed according to hate speech definitions, allowing the model to determine if new content is hateful in a clear, step-by-step process, where each step considers the output of an intermediate step to generate the output. In this way, LLMs are not only capable of executing reasoning-based decisions for identifying online hate speech, but they also offer the advantage of scaling up this process efficiently. Additionally, they can sidestep some of the ethical challenges that social media moderators currently face. In our work, we first formulate a CoT strategy, called HateCoT (\textit{i.e.}, Hate Chain-of-Thought) for reasoning-based decision-making of online hate.

\noindent \textbf{Learning from no or few new samples.} New waves of online hate occur suddenly. To effectively moderate them, we need AI-based discovery techniques that can be updated with no samples or only a few samples so that they can quickly adapt to an updated online hate policy and be deployed. However, training AI models with a few samples is not straightforward, as AI models need large datasets to be sufficiently trained. To effectively moderate new waves of online hate, there is insufficient time to collect and annotate large datasets for training a sufficiently effective classifier. 
\crv{Hence, we require methods that can be rapidly updated to moderate new waves of online hate using only a limited number of new wave samples.}\looseness=-1

\crv{In our work, we concentrate on zero-shot learning through the generation and updating of HateCoT prompts, incorporating new targets and derogatory terms associated with new waves of online hate. This is accomplished by automatically extracting the new targets and derogatory terms using an NLP method.}\looseness=-1

\subsection{Overview of our Framework}

\crv{Our framework, illustrated in Figure~\ref{fig:overview}, comprises three primary components: (1) New Wave Identification; (2) Automatic Prompt Generation and Update; and (3) Automatic New Wave Detection. 
Initially, human moderators from the OSN identify a small set of new wave samples at the start of the new wave. The objective here is to gather a limited dataset, which includes just a few samples of the new wave (like a few tweets), rather than a comprehensive collection of new wave samples.
Subsequently,  HateCoT prompts are generated automatically by identifying new targets and derogatory terms in the seed dataset, verifying their novelty, and updating the HateCoT prompt template with the new targets and derogatory terms. 
A pertinent example is the COVID-19 pandemic, during which Asian Americans were newly targeted with derogatory terms such as ``Wuhan Virus'' and ``Bat Flu.'' These terms were then integrated into the HateCoT prompts, aiding in the detection of emerging online hate trends.
%
Following this, the HateCoT prompts are used to perform reasoning-based decision-making to identify online hate in OSNs. An LLM is leveraged to apply these generated prompts to OSN posts, and the responses from the LLM are then examined to provide answers to the HateCoT prompts. 
%
In the final stage, the responses obtained from the LLM are scrutinized to ascertain whether the posts contain hateful content. Posts identified as hateful are used for extracting targets and derogatory terms and are subsequently flagged for moderation. Additionally, these flagged posts contribute to the automatic expansion of our new wave tweet dataset and facilitate the ongoing refinement of the HateCoT prompts at the \textit{buildup} stage of the new waves.}

\subsection{Our Approach}


\begin{figure*}[t]
\vspace{-0.1cm}
\centering
\resizebox{1.9\columnwidth}{!}{\includegraphics [width=1\linewidth]{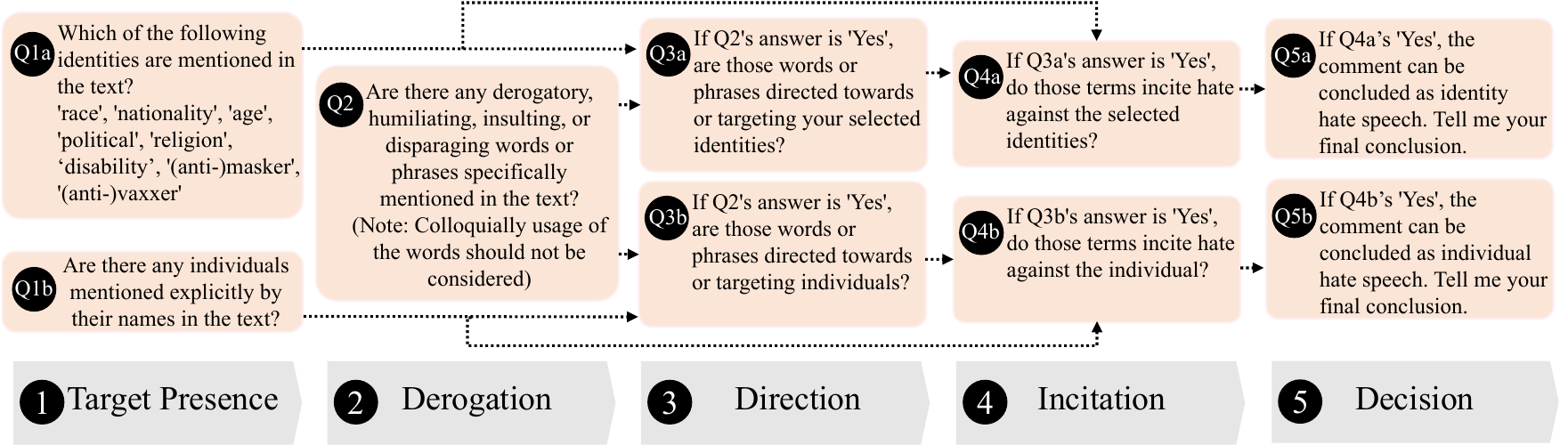}}
\vspace{-0.05cm}
\caption
{HateCoT prompts for new wave detection.}
\vspace{-0.5cm}
\label{fig:cot}
\end{figure*}

\subsubsection{Collecting New Wave Samples}
\crv{A crucial need for addressing new waves is that the detection strategy should be adaptable to the new hate paradigm. One way to achieve this adaptability is by updating the targets and derogatory terms of the new waves. In our framework, we update the HateCoT prompts by continuously retrofitting them with new targets and derogatory terms in new waves.}

\crv{In existing social media platforms like X, human moderators are tasked with monitoring harmful content (already part of their role~\cite{xrules}). Following existing approaches that propose to use human moderators to collect a limited, initial seed dataset of tweets~\cite{paudel2023lambretta}, we anticipate that human moderators can pinpoint a small portion of related tweets as the initial dataset during the \textit{buildup} phase of a new wave, and provide them to \ProjectName{} for the automatic derivation of new targets and derogatory terms.}

\subsubsection{\crv{Automatic Prompt Generation and Update}}
\label{subsubsec:auto_update}
\crv{Based on the seed dataset, \ProjectName{} automatically generates prompts by identifying new targets and derogatory terms, verifying their novelty, and updating a prompt template with these new targets and terms. Then, the prompts are continuously updated by automatically expanding the set of new targets and terms to keep \ProjectName{} up-to-date with the propagation of a new wave.}\looseness=-1
\hfill\\
\textbf{Identifying New Targets and Terms and Verifying Novelty. }\crv{In \ProjectName{}, we use an NLP method to identify new targets and terms, verify their novelty, and automatically expand our dataset of these elements. This approach leverages  KeyBERT~\cite{grootendorst2020keybert} to extract fresh targets and terms from the initial dataset and verifies their novelty using NLTK's WordNet~\cite{farkiya2015natural}. The process iteratively broadens the scope of targets and terms by cross-checking new tweets against our existing dataset, integrating any newly identified elements to ensure comprehensive coverage. }

\crv{
Our NLP method has effectively pinpointed several new targets and terms in posts related to COVID-19. Notably, we uncovered terms such as ``boomers'' (indicating Ageism), ``antimaskers'', and ``antivaxxers'' emerging from the global debates on masks and vaccines, highlighting societal divisions. Our method also discovered derogatory terms like ``BoomerRemover'' (pertaining to ageism) and ``Maskhole'' (targeting anti-maskers).
}

\noindent\textbf{Generating and Updating HateCoT Prompts.} Online hate determination is quite a complex task, which can be broken down into sub-problems that can be individually addressed, and the results put back together to solve the overall problem. We use this characteristic of hate detection to design our HateCoT prompts. Solving this problem by breaking it down into sub-problems significantly improves the ability to perform complex reasoning that hate detection demands. Figure~\ref{fig:cot} depicts HateCoT prompting approach. 


We craft the HateCoT prompts based on the \textit{factors} of identity-based and individual hate, as inferred from their respective definitions. Recall the definition of hate as:

``\textit{Hatred, hostility, or violence} \textit{towards} member(s) of a \textit{race, ethnicity, nation, religion, gender, gender identity, sexual orientation} or any other designated sector of society''~\cite{macavaney2019hate}.\looseness=-1

\crv{From this definition, we identify four key factors of hateful content: (1) Presence of Target, (2) Derogation, (3) Direction, and (4) Incitation. Identifying online hate, therefore, is a comprehensive process that requires assessing each of these factors. We implement this through a CoT approach. This method simplifies the intricate task of detecting online hate by addressing each of the four factors as distinct sub-problems. The outcomes of these sub-problems are then integrated to assess whether content is hateful or not, forming the crucial fifth sub-problem. The specifics of these four factors are detailed below.}

\textit{Presence of Target.} A target should be mentioned in hateful content. In identity-based hate, these targets are based on several identities, such as race, nationality, political affiliation, religion, etc. In hate against individuals, the name or username of the individual is mentioned. In Figure~\ref{fig:cot}, the first sub-problem is based on the presence of a target in the hateful content. This is operationalized with questions Q1a and Q1b, which respectively addressed the presence of identity targets and individual targets.

\textit{Derogation.} From the definition, it can be observed that there is a presence of ``hatred, hostility, or violence'', that is often expressed in textual media using derogatory or disparaging words or phrases. In Figure~\ref{fig:cot}, the second sub-problem is based on the presence of such words or phrases in the content. However, the language in social media platforms also consists of substantial colloquial use of derogatory words such as the f-word, and the sub-problem of detecting derogatory terms must be aware of such colloquial usages. In Figure~\ref{fig:cot}, we operationalize derogation with question Q2, which addresses the mentioning of derogatory terms while being aware of the colloquial use of such terms.

\textit{Direction.} Hate detection is complex enough that the mere presence of targets and derogatory terms are not sufficient to flag a post as hateful. An important factor of hate is that those derogatory terms must be \textit{directed} at the target. For example, the text ``lots of beautiful scenes during the \textit{Chinese} new year, but my \textit{stupid} camera isn't working''. Although a target (Chinese) and a derogatory term (stupid) are mentioned, the term is not directed at the target. The third sub-problem is based on determining whether such terms are directed toward the target. We operationalize this factor with questions Q3a and Q3b in Figure~\ref{fig:cot}, which address the direction toward identities or individuals, respectively. 

\textit{Incitation.} In addition to terms directed at the target, another factor in the detection process is whether the terms incite hate against a target. This differs from the direction of terms towards a target, since benign cases of certain terms directed towards a target can exist, such as ``the f***ing Chinese are winning the space race''. The fourth sub-problem is based on determining whether the detected derogatory terms in the second sub-problem are inciteful of hate toward the detected targets in the first sub-problem. We operationalize this factor with the questions Q4a and Q4b in Figure~\ref{fig:cot}.\looseness=-1

We further define the fifth sub-problem as a \textit{decision}-making task, taking into account the context provided by the answers to all previous sub-problems. This sub-problem concludes the reasoning process and forms the final decision. As shown in Figure~\ref{fig:cot}, this sub-problem is operationalized through the implementation of questions Q5a and Q5b.\looseness=-1

\subsubsection{Automatic New Wave Detection}
\label{subsubsec:entailment-based Few-shot learning}
\hfill\\
\textbf{HateCoT-based New Wave Detection. }
After updating the HateCoT prompts, it's essential to employ a robust model for processing these prompts. Recent advancements have demonstrated that LLMs exhibit enhanced performance in reasoning tasks when prompts are presented in the form of a CoT~\cite{huang2022chain}. The inclusion of intermediate steps in CoT enables the model to engage in more effective thinking and reasoning, thereby significantly improving its decision-making capabilities.

We leverage LLMs to execute our HateCoT prompts, ensuring their design is compatible with various LLMs.  Notably, it has been observed that larger models tend to exhibit superior capabilities in CoT reasoning~\cite {fu2023chain}.

The LLM answers a prompt as follows. Given text input X and prompt t, the final answer is computed as,

\vspace{-0.5mm}
\begin{equation} \label{eq:llmoutput}
    \hat{y} = \text{argmax }p(y|X, t) 
\end{equation}

Instead of asking the LLM the final result $\hat{y}$, we break the problem into many sub-problems as discussed in Section~\ref{subsubsec:auto_update}, such that the model computes $\hat{y} \leftarrow t $ from several intermediate states $\hat{y} \leftarrow t_1 \leftarrow t_2 \dots$. We do this as follows. 

\noindent\textbf{Step 1.} First, we prompt the LLM to output the presence of identity or individual conditioned on the input text, and the identities of the new wave targets, depicted as follows:

\vspace{-0.5mm}
\begin{equation} \label{eq:targespresence}
  \begin{split}
    A1a = \text{argmax }p(a|X, Q1a)\\
    A1b = \text{argmax }p(b|X, Q1b) 
  \end{split}
\end{equation}

In the equation, $a,b, \dots$ are intermediate answers that the LLM could output, such as $Yes$, $No$, and $N/A$. 

\noindent\textbf{Step 2.} Next, we prompt the LLM to compute the presence of derogatory terms based only on the input sentence.
\vspace{-0.5mm}
\begin{equation} \label{eq:derog}
    A2 = \text{argmax }p(c|X, Q2) 
\end{equation}

\noindent\textbf{Step 3.} Then, we prompt the LLM to compute the direction of derogatory terms based on the input sentence and the intermediate outputs from the previous steps.
\vspace{-0.5mm}
\begin{equation} \label{eq:direction}
  \begin{split}
    A3a = \text{argmax }p(d|X, A1a, A2, Q3a)\\
    A3b = \text{argmax }p(e|X, A1b, A2, Q3b) 
  \end{split}
\end{equation}

\noindent\textbf{Step 4.} Next, we prompt the LLM to output whether there is a presence of incitation based on the input sentence and the intermediate outputs from the previous steps.
\vspace{-0.5mm}
\begin{equation} \label{eq:incitation}
  \begin{split}
    A4a = \text{argmax }p(f|X, A1a, A3a, Q4a)\\
    A4b = \text{argmax }p(g|X, A1b, A3b, Q4b) 
  \end{split}
\end{equation}

\noindent\textbf{Step 5.} The final decision is made by prompting the LLM to output a conclusion based on the input sentence and the previous output.
\vspace{-0.5mm}
\begin{equation} \label{eq:incitation}
  \begin{split}
    \hat{y1} = \text{argmax }p(h|X, A4a, Q5a)\\
    \hat{y2} = \text{argmax }p(i|X, A4b, Q5b) 
  \end{split}
\end{equation}

The presence of identity-based hate or individual hate is parsed based on the values of $\hat{y1}$ and $\hat{y2}$, respectively.

        \begin{algorithm}[h!]
    	\caption{HateCoT Reasoning-based Decision-Making Algorithm} 
        \label{alg:css_algorithm}
    	\SetAlgoLined
    	\DontPrintSemicolon
    	\SetNoFillComment
    	$HateCoTPromptTemplate = S$\;
    	{\bfseries Input:} New Waves Dataset (D), Inference Function (I), Large Language Model (M)\;
    	\tcp*{Extract new targets and derogatory terms}
        \For{$x \in D$ }{
    	    $t_x = Targets(x)$ \;
            $d_x = DerogatoryTerms(x)$ \;
    		$NewTargets \cup \{t_x\}$\;
            $NewDerogatoryTerms \cup \{d_x\}$\;
    	}
    		
        
        \For{$s \in HateCoTPromptTemplate$ }{
            $UpdatedHateCoTPrompts = Update(s, NewTargets, NewDerogatoryTerms)$\;
        }
        
        \tcp*{Evaluation of new wave samples}
        
        \For{$n \in D$ }{
            $Decision = I(n, UpdatedCoTPrompts, M)$\;
            \If{Decision = `Yes'}{
            \text{Enforce text control policy.}\\
            \text{Expand Dataset.}
            } \Else{
            \text{Share text.}
            }
            
        }
    \vspace{-1mm}
    \end{algorithm}
\vspace{-3mm}

\noindent\crv{\textbf{Automatic Expansion.}} Lastly, we outline the process for the practical deployment of \ProjectName{} on real-world platforms, like social media, as methodically illustrated in Algorithm~\ref{alg:css_algorithm}. The HateCoT prompt template under typical deployment called $HateCoTPromptTemplate$ is used to control hate speech that the platform is aware of. In the event of a new wave of hate, a small dataset (D) of these new instances is utilized to identify $NewTargets$ and $NewDerogatoryTerms$. Subsequently, the $HateCoTPromptTemplate$ is updated with these new targets and terms, rendering it ready for online deployment. The final step involves running our prompts through an LLM, which then processes the output for monitoring such content on social media platforms. If a new post adheres to the updated hate enforcement policy, it can be flagged. This decision is made by the LLM ($M$) as it evaluates the post against the updated prompts, identifying whether it contains identity hate or individual-targeted hate.

We dynamically expand our dataset of new wave samples by systematically integrating newly detected samples. This expansion is crucial for updating the HateCoT prompts with fresh targets and derogatory terms, as outlined in our NLP method (Section~\ref{subsubsec:auto_update}) and illustrated in Figure~\ref{fig:cot}. Through this ongoing process, we ensure that the HateCoT prompts are automatically refreshed, effectively addressing new-wave instances and guaranteeing comprehensive coverage.\looseness=-1

\begin{figure*}[t]
     \centering
     \begin{subfigure}[t]{0.32\textwidth}
         \centering
         \includegraphics[width=\textwidth]{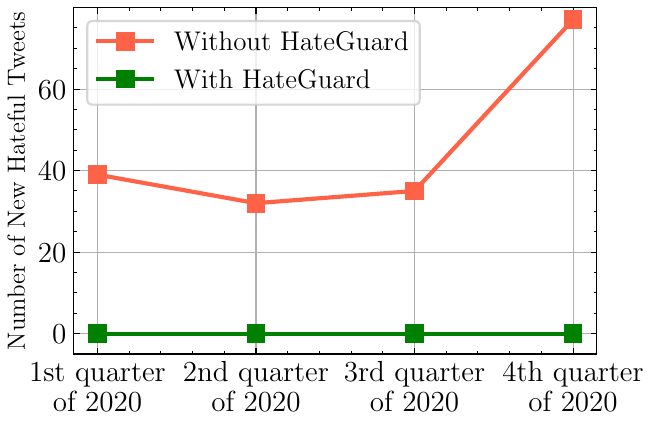}
         \vspace{-5mm}
         \caption{Ageism Hate}
         \label{subfig:age_violaiton}
     \end{subfigure}     
     \begin{subfigure}[t]{0.327\textwidth}
         \centering
         \includegraphics[width=\textwidth]{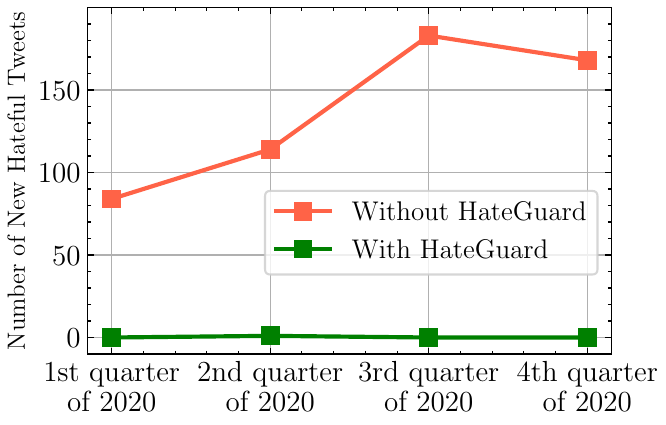}
         \vspace{-5mm}
         \caption{Anti-Asian Hate}
         \label{subfig:asian_violaiton}
     \end{subfigure}
     \begin{subfigure}[t]{0.32\textwidth}
         \centering
         \includegraphics[width=\textwidth]{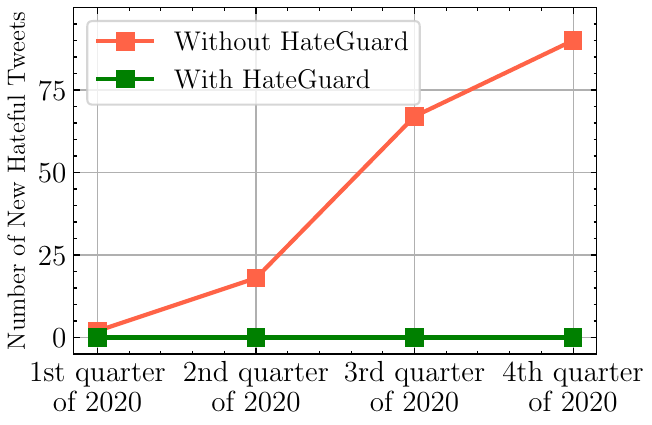}
         \vspace{-5mm}
         \caption{Mask-Related Hate}
         \label{subfig:mask_violation}
     \end{subfigure}
     \begin{subfigure}[t]{0.32\textwidth}
         \centering
         \includegraphics[width=\textwidth]{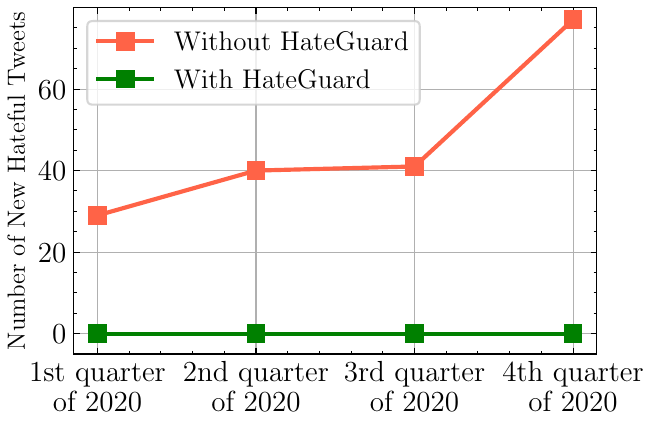}
         \vspace{-5mm}
         \caption{Vaccine-Related Hate}
         \label{subfig:vaccine_violation}
     \end{subfigure}
     \begin{subfigure}[t]{0.327\textwidth}
         \centering
         \includegraphics[width=\textwidth]{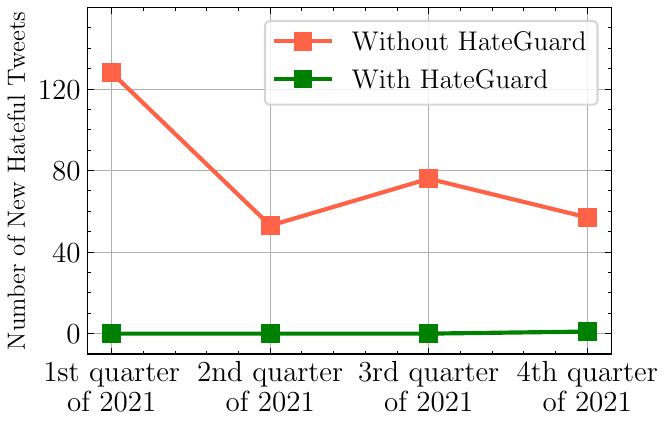}
         \vspace{-5mm}
         \caption{US Capitol Insurrection Hate}
         \label{subfig:us_violation}  
     \end{subfigure}
     \begin{subfigure}[t]{0.32\textwidth}
         \centering
         \includegraphics[width=\textwidth]{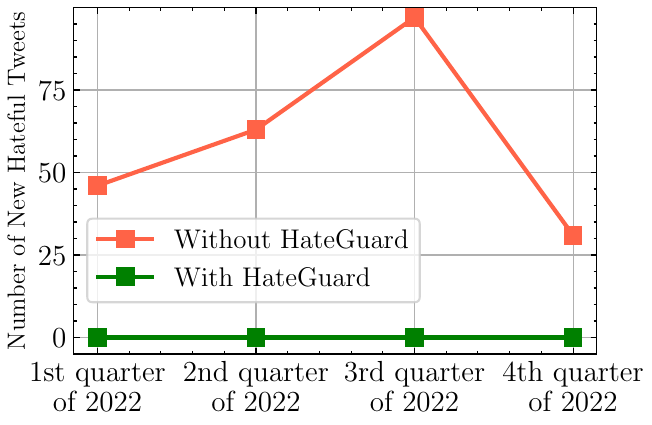}
         \vspace{-5mm}
         \caption{Russian Invasion of Ukraine Hate}
         \label{subfig:rus_violation}  
     \end{subfigure}
    \vspace{-0.04cm} 
    \caption{Deploying \ProjectName{} in 2020 (COVID-19 pandemic), 2021 (US Capitol insurrection), and 2022 (Russian invasion of Ukraine) shows that new wave peaks are significantly reduced (green line). }
    \label{fig:new_hate_violations}
     \vspace{-3mm}
\end{figure*}

\section{Implementation and Evaluation}
\label{sec:evalution}
In this section, we first discuss the implementation of our framework, followed by experiments to evaluate our approaches to all three new waves of online hate from different perspectives. Furthermore, we distinctly focus on the four categories of COVID-19-related online hate (\textit{i.e.}, Anti-Asian, Ageism, Mask, and Vaccine) in addition to the other two new waves in our evaluation, due to the peaks of these categories occurring at different times. Our evaluation goals are summarized below.\looseness=-1



\begin{itemize}
    \item Understanding the effectiveness of \ProjectName{} by investigating the number of policy violations per quarter in the years 2020 (COVID-19 pandemic), 2021 (US Capitol insurrection), and 2022 (Russian invasion of Ukraine) ($\S$~~\ref{subsec:quater_violation}).
    
    \item Evaluating the effectiveness of our framework by comparing it with existing benchmarks ($\S$~~\ref{subsec:comparisonbenchmarks}).

     \item Analyzing the effectiveness of \ProjectName{} by comparing it with state-of-the-art ZSL, FSL, and generalized prompt strategies. ($\S$~~\ref{subsec:comparisonotherzslfsl}).

    
    
    \item Running \ProjectName{} on ``in-the-wild'' unlabeled samples in our dataset. ($\S$~~\ref{subsec:unlabelledsamples}).
\end{itemize}
\subsection{Implementation}
\label{subsec:implementation}

We used the GPT-4 model from the official OpenAI API endpoints to run HateCoT prompts as well as the generalized prompts~\cite{li2023hot}. \crv{We used KeyBERT version 0.8.3 and NLTK version 3.8.1 in our NLP method.}
Our labeled dataset was primarily utilized for the majority of our tests, while the unlabeled dataset was employed for the studies in Section~\ref{subsec:nateuofnewwaves} and the evaluations in Section~\ref{subsec:unlabelledsamples}. Furthermore, we provide detailed discussions on specific parameter settings, excluding those set to defaults, in the respective evaluation sections. 

\begin{table*}[]
\centering
\resizebox{1\textwidth}{!}{%
\begin{tabular}{cccccccccccccccccc}
\toprule
\multirow{3}{*}{\textbf{Wave Type}}                                                           & \multirow{3}{*}{\textbf{Method}} & \multicolumn{4}{c}{\textbf{Quarter 1 (Jan-Mar)}}                                                                    & \multicolumn{4}{c}{\textbf{Quarter 2 (Apr-Jun)}}                                                                       & \multicolumn{4}{c}{\textbf{Quarter 3 (Jul-Sep})}                                                                    & \multicolumn{4}{c}{\textbf{Quarter 4 (Oct-Dec)}}                                                                       \\
\cmidrule(r){3-6} \cmidrule(r){7-10} \cmidrule(r){11-14} \cmidrule(r){15-18}     
 & & \textbf{\begin{tabular}[c]{@{}c@{}}\# of\\ Tweets\end{tabular}} & \textbf{\begin{tabular}[c]{@{}c@{}}Acc-\\ uracy\end{tabular}} & \textbf{\begin{tabular}[c]{@{}c@{}}Prec-\\ ision\end{tabular}} & \textbf{\begin{tabular}[c]{@{}c@{}}Rec-\\ all\end{tabular}} & \textbf{\begin{tabular}[c]{@{}c@{}}\# of\\ Tweets\end{tabular}} & \textbf{\begin{tabular}[c]{@{}c@{}}Acc-\\ uracy\end{tabular}} & \textbf{\begin{tabular}[c]{@{}c@{}}Prec-\\ ision\end{tabular}} & \textbf{\begin{tabular}[c]{@{}c@{}}Rec-\\ all\end{tabular}} & \textbf{\begin{tabular}[c]{@{}c@{}}\# of\\ Tweets\end{tabular}} & \textbf{\begin{tabular}[c]{@{}c@{}}Acc-\\ uracy\end{tabular}} & \textbf{\begin{tabular}[c]{@{}c@{}}Prec-\\ ision\end{tabular}} & \textbf{\begin{tabular}[c]{@{}c@{}}Rec-\\ all\end{tabular}} & \textbf{\begin{tabular}[c]{@{}c@{}}\# of\\ Tweets\end{tabular}} & \textbf{\begin{tabular}[c]{@{}c@{}}Acc-\\ uracy\end{tabular}} & \textbf{\begin{tabular}[c]{@{}c@{}}Prec-\\ ision\end{tabular}} & \textbf{\begin{tabular}[c]{@{}c@{}}Rec-\\ all\end{tabular}} \\
\midrule
\multicolumn{18}{c}{\textbf{- Overall Results -}}\\
\midrule
\multirow{3}{*}{
\begin{tabular}[c]{@{}c@{}}Total\\ (2020-2022)\end{tabular}
} 
& \ProjectNameTable{} & 
\multirow{3}{*}{928} & \textbf{0.95} & \textbf{0.95}    & \textbf{0.94} & 
\multirow{3}{*}{893} & \textbf{0.94} & \textbf{0.94} & \textbf{0.93}    
& \multirow{3}{*}{1148} & \textbf{0.94} & \textbf{0.94} & \textbf{0.93} & 
\multirow{3}{*}{1031} & \textbf{0.94}    & \textbf{0.95}    & \textbf{0.92}    \\
& BERT-base & & 0.74 & 0.81 & 0.34& & 0.82& 0.76& 0.71& &0.84 & 0.82& 0.79& & 0.83& 0.86&0.8 \\
& Tweet-NLP & & 0.7 & 0.73 &0.23 & & 0.83&0.79 &0.77 & &0.84 & 0.83& 0.8& &0.83 &0.84 &0.8 \\
\midrule
\multicolumn{18}{c}{\textbf{- Category-wise Results -}}\\
\midrule
\multirow{3}{*}{\begin{tabular}[c]{@{}c@{}}Ageism\\ (2020)\end{tabular}}
& \ProjectNameTable{} & \multirow{3}{*}{186} & \textbf{0.94} & \textbf{0.91} & \textbf{0.92} & \multirow{3}{*}{117} & \textbf{0.95}    & \textbf{0.95}    & \textbf{0.95}    & \multirow{3}{*}{114} & \textbf{0.95}    & \textbf{0.95} & \textbf{0.95} & \multirow{3}{*}{161} & \textbf{0.95}    & \textbf{0.94}    & \textbf{0.96} \\
 & BERT-base               &                                                        & 0.82          & 0.6           & 0.44       &                                                        & 0.8           & 0.68          & 0.53          &                                                        & 0.79          & 0.68          & 0.6        &                                                        & 0.74          & 0.72          & 0.76          \\
 & Tweet-NLP               &                                                        & 0.79          & 0.5           & 0.15       &                                                        & 0.87          & 0.79          & 0.72          &                                                        & 0.86          & 0.74          & 0.83       &                                                        & 0.72          & 0.79          & 0.57          \\
\multirow{3}{*}{\begin{tabular}[c]{@{}c@{}}Asian\\ (2020)\end{tabular}}
& \ProjectNameTable{} & \multirow{3}{*}{179} & \textbf{0.96} & \textbf{0.96} & \textbf{0.97} & \multirow{3}{*}{296} & \textbf{0.93} & \textbf{0.93} & \textbf{0.93} & \multirow{3}{*}{331} & \textbf{0.94} & \textbf{0.95} & \textbf{0.94} & \multirow{3}{*}{262} & \textbf{0.95} & \textbf{0.94} & \textbf{0.98}    \\
 & BERT-base               &                                                        & 0.68          & 0.91          & 0.35       &                                                        & 0.84          & 0.79          & 0.8           &                                                        & 0.85          & 0.86          & 0.87       &                                                        & 0.87          & 0.88          & 0.92          \\
                                                                                    & Tweet-NLP               &                                                        & 0.63          & 0.77          & 0.29       &                                                        & 0.84          & 0.84          & 0.72          &                                                        & 0.84          & 0.86          & 0.84       &                                                        & 0.85          & 0.91          & 0.86          \\
\multirow{3}{*}{\begin{tabular}[c]{@{}c@{}}Mask\\ (2020)\end{tabular}}
& \ProjectNameTable{} & \multirow{3}{*}{16} & \textbf{0.99} & \textbf{0.99} & \textbf{0.99} & \multirow{3}{*}{64} & \textbf{0.94}    & \textbf{0.96} & \textbf{0.88} & \multirow{3}{*}{249} & \textbf{0.98} & \textbf{0.95} & \textbf{0.97} & \multirow{3}{*}{199} & \textbf{0.96} & \textbf{0.97} & \textbf{0.94}    \\
                                                                                    & BERT-base               &                                                        & 0.75          & 0             & 0          &                                                        & 0.79          & 0.78          & 0.39          &                                                        & 0.85          & 0.75          & 0.66       &                                                        & 0.8           & 0.75          & 0.86          \\
                                                                                    & Tweet-NLP               &                                                        & 0.94          & 0.67          & \textbf{0.99} &                                                        & 0.86          & 0.85          & 0.61          &                                                        & 0.87          & 0.75          & 0.78       &                                                        & 0.84          & 0.8           & 0.88          \\
\multirow{3}{*}{\begin{tabular}[c]{@{}c@{}}Vaccine\\ (2020)\end{tabular}}
& \ProjectNameTable{} & \multirow{3}{*}{78} & \textbf{0.98} & \textbf{0.99} & \textbf{0.96} & \multirow{3}{*}{114} & \textbf{0.92} & \textbf{0.9} & \textbf{0.92}    & \multirow{3}{*}{104} & \textbf{0.93} & \textbf{0.93} & \textbf{0.91} & \multirow{3}{*}{226} & \textbf{0.94} & \textbf{0.95} & \textbf{0.92}    \\

& BERT-base               &                                                        & 0.76          & 0.92          & 0.38       &                                                        & 0.78          & 0.68          & 0.7           &                                                        & 0.85          & 0.79          & 0.83       &                                                        & 0.84          & 0.75          & 0.79          \\
                                                                                    & Tweet-NLP               &                                                        & 0.72          & 0.77          & 0.35       &                                                        & 0.75          & 0.6           & 0.83          &                                                        & 0.8           & 0.75          & 0.73       &                                                        & 0.88          & 0.83          & 0.82          \\
\multirow{3}{*}{\begin{tabular}[c]{@{}c@{}}US Capitol\\ (2021)\end{tabular}}
& \ProjectNameTable{} & \multirow{3}{*}{311} & \textbf{0.91} & \textbf{0.91} & \textbf{0.88} & \multirow{3}{*}{112} & \textbf{0.99}    & \textbf{0.97} & \textbf{0.99} & \multirow{3}{*}{158} & \textbf{0.9}    & \textbf{0.89}    & \textbf{0.9} & \multirow{3}{*}{123} & \textbf{0.9} & \textbf{0.9}    & \textbf{0.9} \\
 & BERT-base               &                                                        & 0.68          & 0.79          & 0.31       &                                                        & 0.85          & 0.85          & 0.83          &                                                        & 0.82          & 0.85          & 0.76       &                                                        & 0.84          & 0.78          & 0.89          \\
                                                                                    & Tweet-NLP               &                                                        & 0.63          & 0.7           & 0.16       &                                                        & 0.83          & 0.79          & 0.87          &                                                        & 0.82          & 0.87          & 0.72       &                                                        & 0.78          & 0.75          & 0.79          \\
\multirow{3}{*}{\begin{tabular}[c]{@{}c@{}}Russia\\ -Ukraine\\ (2022)\end{tabular}} & 
\ProjectNameTable{} & \multirow{3}{*}{158} & \textbf{0.95} & \textbf{0.95} & \textbf{0.93} & \multirow{3}{*}{190} & \textbf{0.94} & \textbf{0.95} & \textbf{0.92} & \multirow{3}{*}{192} & \textbf{0.94} & \textbf{0.95} & \textbf{0.93} & \multirow{3}{*}{60} & \textbf{0.9}    & \textbf{0.92}    & \textbf{0.9}    \\
 & BERT-base               &                                                        & 0.8           & 0.85          & 0.37       &                                                        & 0.8           & 0.72          & 0.62          &                                                        & 0.82          & 0.82          & 0.81       &                                                        & 0.83          & 0.82          & 0.87          \\
                                                                                    & Tweet-NLP               &                                                        & 0.77          & 0.92          & 0.24       &                                                        & 0.84          & 0.82          & 0.65          &                                                        & 0.86          & 0.88          & 0.84       &                                                        & 0.85          & 0.89          & 0.81         
\\

\bottomrule
\end{tabular}%
}
\caption{Comparing \ProjectNameTable{}  against the existing  benchmarks.}
\vspace{-5mm}
\label{tab:violations_compare_baseline_new}
\end{table*}

\subsection{\NishNDSS{Baselines}}
To evaluate our framework, we present several key baselines for comparison: (1) two existing approaches, BERT-base~\cite{mathew2021hatexplain} and Tweet-NLP~\cite{barbieri2020tweeteval}, as discussed in Section~\ref{subsec:quater_violation}; (2) fundamental approaches such as FSL~\cite{huggingface} and Meta-EFL~\cite{wang2021entailment},  covered in Section~\ref{subsec:comparisonotherzslfsl}; and (3) a generalized prompt strategy for hate speech detection, derived from existing literature~\cite{li2023hot}, which is elaborated in Section~\ref{subsec:comparisonotherzslfsl}.

\subsection{Effectiveness of \ProjectName{} in Reducing Number of Violations}
\label{subsec:quater_violation}
In this section, we investigated the effectiveness of \ProjectName{} in efficiently reducing the peaks of new waves of online hate. We were especially interested in learning whether \ProjectName{} can reduce the peaks of new hate waves in a \emph{real-world deployment simulation} scenario. 
We first categorized all the tweets belonging to the three new waves of online hate samples in the dataset according to different quarters and types. For COVID-19-related hate, we focused on the four quarters of 2020 since that year of the pandemic witnessed numerous waves of online hate. On a similar basis, we focused on the four quarters of 2021 for the US Capitol insurrection-related hate and 2022 for the Russian invasion of Ukraine-related hate. For these same periods, we deployed \ProjectName{} for different hate categories and recorded the number of violations with our framework.\looseness=-1

Figure~\ref{fig:new_hate_violations} depicts the temporal progress of the spread of new waves without and with \ProjectName{}, respectively, wherein the red line in Figure~\ref{fig:new_hate_violations} indicates how many violations were made in each quarter, and the green line indicates the number of violations after deploying \ProjectName{}. 
From Figure~\ref{fig:new_hate_violations}, we observed that all three new waves~(including the four categories of COVID-19-related hate) reached significant peaks in at least one of the four quarters~(for instance, the third quarter for Anti-Asian hate and the fourth quarter for Vaccine-related hate). 
However, the peaks were significantly reduced in those specific quarters, and the overall violations were reduced in each of the quarters with the deployment of our framework. In most quarters, \ProjectName{} completely stopped new waves from occurring. For example, the peaks of each category of new waves were effectively reduced, demonstrating that our framework is capable of moderating various types of new waves by only updating the prompts.\looseness=-1

\subsection{Comparison with Existing Benchmarks}
\label{subsec:comparisonbenchmarks}
\vspace{-0.4cm}
In this experiment, we studied the effectiveness of \ProjectName{} in discovering new waves of online hate in comparison to existing benchmarks of transformer models~\cite{vaswani2017attention}.  We used two transformer models as benchmarks. The first, BERT-base-uncased mode, was fine-tuned using a leading dataset for hateful speech~\cite{mathew2021hatexplain}. The second, Tweet-NLP model~\cite{barbieri2020tweeteval}, is widely recognized for its effectiveness in hate speech detection on X, which is particularly relevant as our dataset of new hate waves was sourced from the same platform. For both models, we utilized the official implementations available online~\cite{huggingface}.
To clearly illustrate the effectiveness of \ProjectName{} in addressing the emergence of new waves of online hate, we conducted a comparative analysis between our framework and these baseline models. This evaluation was performed quarterly, focusing on metrics such as accuracy, precision, and recall. 
In each quarter, we incrementally fine-tuned the benchmark models with the data from the preceding quarter. For example, in Q1, the models were applied in their original form. In Q2, we incorporated data from Q1 for further training, and this process continued sequentially. 
This approach was designed to mimic a real-world OSN scenario, where models are periodically updated to reflect new trends and policies in online hate speech. By following the guidelines provided in the referenced studies~\cite{mathew2021hatexplain}~\cite{barbieri2020tweeteval}, we trained both models over 50 training epochs. 
To evaluate \ProjectName{}, we first randomly selected seed data with 10 to 20 tweets from the first month of each quarter to identify new targets and derogatory terms. We utilized data from the remaining months of each quarter to assess \ProjectName{}’s performance.


The outcomes of our experiment are presented in Table~\ref{tab:violations_compare_baseline_new}. In the summary of Overall Results, \ProjectName{}  significantly surpassed the existing benchmarks, demonstrating a remarkable detection rate in identifying emerging new waves of online hate throughout all quarters. This superior performance is highlighted by the bolded numbers in Table~\ref{tab:violations_compare_baseline_new}.
Specifically, \ProjectName{} achieved impressive F1 scores of 0.95 in Q1, 0.94 in Q2, 0.94 in Q3, and 0.94 in Q4, indicating that it was effective and did not overfit the training data. 
\crv{Upon further investigation into each type of new wave, the results (in ``Category-wise Results'') consistently revealed that \ProjectName{} was markedly more effective in identifying and flagging the content associated with these new waves compared to benchmark methods.}
For instance, \ProjectName{} achieved a precision of 0.91 and a recall of 0.92 on Asian hate, a major category of concern during COVID-19 in Q1, maintaining this high level of performance through to Q4.
Furthermore, \ProjectName{} exhibited exceptional accuracy in its performance, achieving a 91\% accuracy rate in Q1, the peak period for US Capitol-related hate, and maintaining this high accuracy level at 94\% in Q3, the peak quarter for Russian-Ukraine-related hate. 
Interestingly, our framework demonstrated notable improvement in the first quarter, even with a small number of training samples. This suggests it can effectively moderate with minimal data, akin to zero-shot learning. As we approached the final quarter, the baseline models began to catch up slightly. For example, the BERT-base-uncased model achieved a 76\% recall rate for Ageism, and the Tweet-NLP model reached an 88\% recall rate for Mask-related hate, indicating their increasing effectiveness as more data becomes available.
While the baseline models may be adequate for identifying hate speech within larger datasets, our framework demonstrated a significantly greater capability in effectively moderating new waves.\looseness=-1

\subsection{Comparison with other ZSL and FSL Methods}
\label{subsec:comparisonotherzslfsl}

\begin{table}[!b]
\vspace{-5mm}
\centering
\resizebox{0.8\columnwidth}{!}{%
\begin{tabular}{ccccc}

\toprule
\textbf{Wave Type}                                                         & \textbf{Method}                                          & \textbf{Accuracy} & \textbf{Precision} & \textbf{Recall} \\
\midrule
\multirow{4}{*}{Ageism}                                                    & ZSL                                                      & 0.69              & 0.51               & 0.33            \\
                                                                           & FSL                                                      & 0.5               & 0.36               & 0.52            \\
                                                                           & GP & 0.84              & 0.69               & 0.88            \\
                                                                           & \ProjectNameTable{}                                                      & \textbf{0.95}     & \textbf{0.94}      & \textbf{0.94}      \\
\multirow{4}{*}{Asian}                                                     & ZSL                                                      & 0.74              & 0.75               & 0.75            \\
                                                                           & FSL                                                      & 0.5               & 0.53               & 0.47            \\
                                                                           & GP & 0.87              & 0.82               & \textbf{0.96 }           \\
                                                                           & \ProjectNameTable{}                                                      & \textbf{0.95}              & \textbf{0.95}               & 0.94            \\
\multirow{4}{*}{Mask}                                                      & ZSL                                                      & 0.74              & 0.65               & 0.51            \\
                                                                           & FSL                                                      & 0.49              & 0.34               & 0.48            \\
                                                                           & GP & 0.84              & 0.74               & 0.8             \\
                                                                           & \ProjectNameTable{}                                                      & \textbf{0.96}     & \textbf{0.96}      & \textbf{0.93}      \\
\multirow{4}{*}{Vaccine}                                                   & ZSL                                                      & 0.73              & 0.72               & 0.4             \\
                                                                           & FSL                                                      & 0.5               & 0.37               & 0.48            \\
                                                                           & GP & 0.85              & 0.73               & 0.9             \\
                                                                           & \ProjectNameTable{}                                                      & \textbf{0.94}              & \textbf{0.94}               & \textbf{0.93}               \\
\multirow{4}{*}{US Capitol}                                                & ZSL                                                      & 0.69              & 0.79               & 0.42            \\
                                                                           & FSL                                                      & 0.54              & 0.49               & 0.56            \\
                                                                           & GP & 0.83              & 0.75               & 0.9             \\
                                                                           & \ProjectNameTable{}                                                      & \textbf{0.92}     & \textbf{0.92}      & \textbf{0.92}      \\
\multirow{4}{*}{\begin{tabular}[c]{@{}c@{}}Russia\\ -Ukraine\end{tabular}} & ZSL                                                      & 0.81              & 0.84               & 0.63            \\
                                                                           & FSL                                                      & 0.49              & 0.4                & 0.49            \\
                                                                           & GP & 0.87              & 0.79               & 0.93            \\
                                                                           & \ProjectNameTable{}                                                      & \textbf{0.94}     & \textbf{0.95}      & \textbf{0.94}     \\
\bottomrule
\end{tabular}
}
\vspace{-1mm}
\caption{\crv{Results of different ZSL and FSL methods.}}
\label{tab:fsl}
\vspace{-6mm}
\end{table}

\begin{figure*}[t]
\vspace{-0.1cm}
\centering
\resizebox{1.8\columnwidth}{!}{\includegraphics [width=1\linewidth]{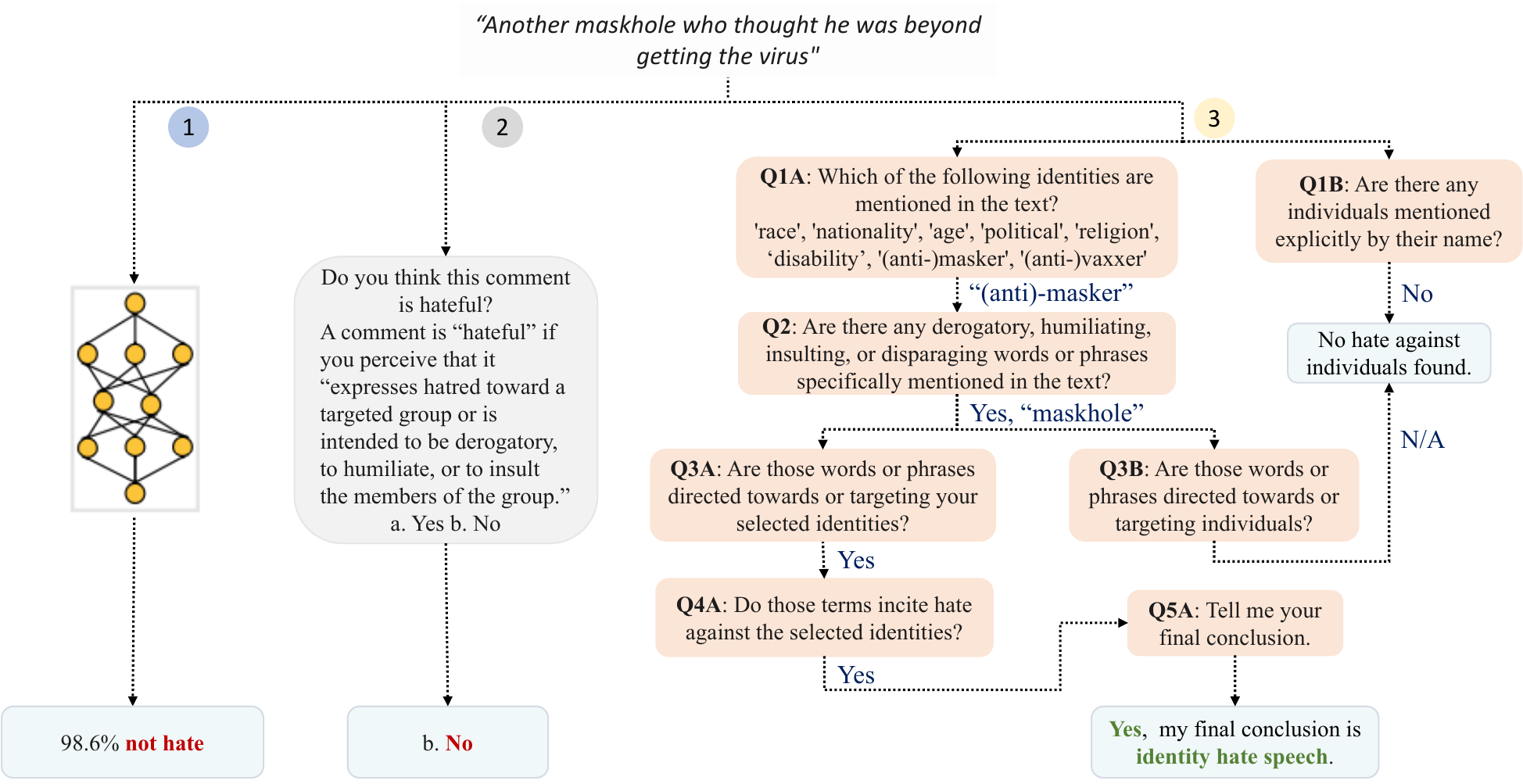}}
\caption
{\protect\inlinegraphics{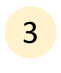}HateCoT's prompt-based reasoning for new wave decision-making compared to \protect\inlinegraphics{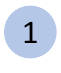}RoBERTa hate speech detection model~\cite{vidgen2021learning} and \protect\inlinegraphics{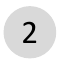}general prompting method~\cite{li2023hot}.}
\vspace{-0.5cm}
\label{fig:decisiontree}
\end{figure*}

To study the importance of our HateCoT update strategy based on the ZSL paradigm, we compared \ProjectName{} with one few-shot learning model, one zero-shot model, and a zero-shot model based on general prompting. The first method is a RoBERTa hate speech detection model fine-tuned by a benchmark dataset~\cite{vidgen2021learning} (referred to as ``ZSL'' in this evaluation), the second method is the Meta-EFL model proposed by Meta~\cite{wang2021entailment} (referred to as ``FSL'' in this evaluation), and the third method is a general prompting method~\cite{li2023hot} based on ZSL (referred to as ``GP''). 
For the first model, ZSL, we used the model by directly running it on the samples in our dataset.
For the second method, we set the hypothesis for Meta-EFL as outlined in the original paper~\cite{wang2021entailment}, where the entailment hypothesis is defined as ``This tweet contains hate speech''.  In this experiment, we trained all the models for the same number of epochs and set the parameters as mentioned in their original fine-tuned state. For GP, we used the prompt mentioned in the original paper~\cite{li2023hot}.\looseness=-1

Table~\ref{tab:fsl} presents the results of this experiment, depicting a detailed comparison of the accuracy, precision, and recall metrics of these methods with our framework. 
Notably, our framework outperformed the other three models based on ZSL/FSL. The primary aim of this experiment was to assess the effectiveness of the HateCoT prompts and our update strategy. First, we note that our approach significantly outperformed the ZSL and FSL models by a large margin, which could be attributed to the lack of any contextual detection offered by these models. While the GP strategy performed better than the other models, our prompting strategy vastly outperformed GP, especially in terms of precision. These results shed light on the contribution of chain-of-thought reasoning in the domain of hateful content detection, especially in new waves. 
To further clarify our approach, we display a specific example in Figure~\ref{fig:decisiontree}, featuring the sentence “Another maskhole who thought he was beyond getting the virus,” analyzed using the ZSL model, the GP method, and our approach. This comparison highlights how our model's effectiveness is enhanced by its responses to a series of chained prompts.
By breaking down the problem into a decision-making task involving sub-problems, such as identifying identities, derogatory terms, and their intended direction, the LLM is prompted to consider a range of possibilities, a crucial aspect in hate speech detection. This methodical, reasoning-based examination of a sample leads to an improvement in both precision and recall, as the final decision is derived through these intermediate steps rather than an immediate classification. More examples demonstrating our approach are available for review in Table~\ref{tab:moresamples}, Appendix~\ref{app:samples}.

\subsection{\crv{Running \ProjectName{} on ``In-the-Wild'' Samples}}
\label{subsec:unlabelledsamples}
In this study, we tested our approach on unlabeled samples from our dataset to create a real-world, ``in-the-wild'' scenario. The goal of the experiment was to spontaneously flag potential hate content using \ProjectName{}, and do a post-fact verification on whether the flagged content is indeed hateful. \looseness=-1

In our experiment, we first randomly sampled 1,500 unlabeled samples from our dataset and ran \ProjectName{} on these random samples. Two experts who are authors of this paper independently labeled the 1,500 samples based on the code book developed for labeling the new waves of online hate. We considered the two expert's annotations as ground truth and then analyzed the performance of \ProjectName{} on the same samples. \ProjectName{} was successfully able to flag 100\% of the samples labeled as hate in the random in-the-wild samples, which indicates that it could be deployed as a main defense strategy on posts in real-world applications. Furthermore, in this experiment, \ProjectName{} achieved a precision of 0.97 and a recall of 0.99, indicating that it is capable of achieving a very low False Positives (FP) rate. We hold the belief that \ProjectName{} could be a potentially potent defense mechanism in times of crisis, such as elections, where unchecked new waves proliferate. In such cases, it remains crucial to maintain the balance between curtailing hate speech and safeguarding free speech and legitimate criticisms, which are integral to the foundation of democratic societies. Given the notably low FP rate achieved by our approach, we hold confidence that it can effectively identify hateful posts while avoiding inadvertent censorship of harmless content. This, in turn, contributes to alleviating the burden and exhaustion endured by human social media moderators in the task of hate content moderation.

\crv{Additionally, to test the efficacy of our NLP method (Section~\ref{subsubsec:auto_update}), we conducted an assessment, utilizing 10 tweets from the COVID-19 pandemic dataset at the beginning of the buildup phase of Mask-hate. Our method effectively identified 1 target (``antimaskers'') and 2 terms (``maskhole'' and ``maskoff'') about Mask-hate based on the initial dataset, expanding to 39 targets/terms by the end of the buildup phase of Mask-hate. Our analysis indicated a strong alignment between the new targets and terms identified through our NLP method and those gleaned from the authors' manual analysis.}

\vspace{-0.1cm}
\section{Discussion } 
\label{sec:discussion}
\vspace{-0.2cm}

\noindent \textbf{Broader Impacts of Chain-of-Thought Reasoning on LLM-based Security Applications.} To the best of our knowledge, our work is the first to leverage CoT reasoning within LLMs to address a critical cybersecurity challenge. We advocate the CoT strategy for scenarios requiring intricate decision-making, as many cybersecurity applications based on LLMs could significantly benefit from this approach, as evidenced in our study. A case in point is Microsoft’s introduction of the Security Copilot~\cite{microsoftsecuritycopilot}, a cutting-edge tool that integrates ChatGPT for sophisticated threat analysis. While our current focus is on new waves of online hate, the versatility of our methodology is evident in its potential applicability to analogous challenges~\cite{mahdavifar2019application,manocchio2023flowtransformer,souri2018state}. For instance, our approach could expediently enhance LLM-based malware detection systems~\cite{llm_malware}, enabling them to swiftly adapt to zero-day malware with minimal initial samples. CoT-based reasoning might also streamline the training processes for transformer-based network intrusion detection systems~\cite{manocchio2023flowtransformer}, minimizing their dependency on extensive data. Additionally, CoT could be instrumental in generating varied fuzzing inputs across multiple programming languages~\cite{xia2023universal}. We encourage cybersecurity professionals and researchers to further investigate the application of CoT principles in cybersecurity solutions.\looseness=-1

\noindent \textbf{Limitation.} 
In our study, we primarily focus on text modality, the most common medium for disseminating online hate. However, emerging research highlights the increasing use of other modalities like images~\cite{kiela2019supervised}, videos~\cite{ottoni2018analyzing,yang2021youtubers}, and speech~\cite{sullivan2019platforms} in propagating online hate. Incorporating these modalities into our evaluations could offer a more comprehensive assessment of our framework. Additionally, our current analysis is confined to English-language posts. Expanding our scope to include other languages would provide a deeper insight into the effectiveness of our framework in diverse linguistic contexts. Moreover, we envision the application of \ProjectName{} by content moderators, especially during critical crisis events, to identify and manage online hate content. However, there is a potential risk of unintended consequences, such as misclassifying benign comments as hate speech, if \ProjectName{} is employed without proper supervision or review.

\noindent \textbf{Ethical Considerations.} We used workers from AMT to annotate the new waves of online hate dataset. Our data collection task was approved by IRB. We also warned workers about potential hateful content before they agreed to work on our task. In our paper, we have taken steps to minimize depicting samples of hate from our dataset, and we have carefully censored words that are extremely hateful or derogatory. We ensured the removal of mentions to user accounts so that user accounts could not be traced via public social media. \looseness=-1

\vspace{-0.1cm}
\section{Conclusion and Future Work}
\vspace{-0.2cm}
\label{sec:conclusion}
In this work, we conducted a large-scale experiment to study the nature of new waves of online hate and showed how a new wave of online hate reaches a peak of activity, during which online hate spreads unabated. Then, we examined the capabilities of the existing moderation tools and found that they are significantly limited when used against the new waves of online hate. 
We proposed a novel framework \ProjectName{} to practically address the problem of new waves of online hate. 
Our evaluation shows that \ProjectName{} can significantly reduce the number of violations caused by new wave samples, and help in practically addressing this critical problem. \looseness=-1


\crv{In the future, we aim to expand our framework to accommodate multilingual scenarios. This is crucial given the diverse linguistic landscape of global OSN platforms. To realize this, we plan to investigate multilingual encoders~\cite{dabre2020survey} that seamlessly integrate into our framework, primarily focusing on updating the content encoding step. Additionally, we plan to broaden our framework to include multimodal scenarios, particularly those combining image and text modalities, such as memes. Given the recent rise of memes as a medium for propagating online hate~\cite{kiela2019supervised}, we will explore how CoT prompting can be used to control new waves of multimodal online hate. Additionally, we plan to investigate the efficacy of alternative reasoning approaches, such as tree-of-thought (ToT)~\cite{yao2023tree} and graph-of-thought (GoT)~\cite{besta2023graph}, in combating online hate. Finally, while our current work involves manually crafted prompts, we are keen on examining other prompt engineering techniques~\cite{liu2023pre} that could further refine and enhance the efficacy of our prompts.}



\vspace{-0.1cm}
\section{Acknowledgements}
\vspace{-0.2cm}
This material is based upon work supported in part by the National Science Foundation (NSF) under Grant No. 2245983, 2129164, 2114982, 2228617, 2120369, 2237238, 2239605, 2228616, and 2114920, and a National Centers of Academic Excellence in Cybersecurity grant No. H98230-22-1-0307. We thank Dr. Bimal Viswanath from Virginia Tech for his valuable suggestions on enhancing the evaluation of \ProjectName{}.\looseness=-1

\bibliographystyle{IEEEtran}
\bibliography{bibtexes}


\begin{appendices}
\section{}
\label{subsec:hashtags_list}
We provide the complete list of hashtags used for data collection in this work in Table~\ref{tab:list_of_hashtags}.

\begin{table}[h]
\centering
\resizebox{\columnwidth}{!}{%
\setlength\tabcolsep{0.2ex}
\begin{tabular}{p{2.4cm} p{7.5cm}}
\toprule
\textbf{Category} & \textbf{Hashtags} \\
\midrule
Anti-Asian        & `\#chinesevirus', `\#chinavirus', `\#wuhanflu', `\#batmaneatingflu', `\#yellowmanflu', `\#fuckchina', `\#bombchina', `\#ChinaLiedPeopleDied', `boycottchina' \\
Ageism            & `\#boomerremover', `\#boomerentomber', `\#okboomer', `\#boomerdeath', `\#oldaf', `\#boomermoober' \\
Mask              & `\#NoMask', `\#NoMasks', `\#MasksOff', `\#MasksDontWork', `\#WearAMask', `\#WearADamnMask', `\#MaskUp' \\ 
Vaccine           & `\#covid19vaccine', `\#covidvaccine', `\#pfizercovidvaccine', `\#modernacovidvaccine', `\#astrazenecacovidvaccine', `\#biontechcovidvaccine', `\#covidiots', `\#iwillnotcomply' \\ 
US Capitol Insurrection & `\#MAGARioters', `\#MAGAMorons', `\#MAGATerrorists', `\#TrumpCrimeFamily', `\#TrumpIsALaughingStock', `\#TrumpCrimeSyndicate', `\#TrumpInsurrection', `\#TrumpCrimeFamilyForPrison' \\
Russian Invasion of Ukraine & `\#NaziRussia', `\#FkPutin', `\#Rushit', `\#getoutrussia', `\#BanRussia', `\#RussiaUkraineConflict', `\#RussiaIsATerroristState', `\#SanctionRussiaNow', `\#PutinWar', `\#Zelenskyclown', `\#ZelenskyJoker', `\#KillPutin'
\\
\bottomrule
\end{tabular}
}
\caption{List of hashtags.}
\label{tab:list_of_hashtags}
\vspace{-5mm}
\end{table}

\section{}
\label{app:mturkannotations}
We introduce our code book, devised by two experts and utilized for the Mturk annotation task in Table~\ref{tab:codebook}.

\begin{table}[h!]
\center
\resizebox{\columnwidth}{!}{
\begin{tabular}{l}
\toprule
\begin{tabular}[c]{@{}l@{}}Please read the short instruction, then answer questions according to the following sentences:\\ \$\{text\}\end{tabular}                                                                                                                             \\
\midrule 
\begin{tabular}[c]{@{}l@{}}Q1A: Which of the following identities are mentioned in the text? (select one or multiple\\ options, ``No identity'' if there's not)\\
$\square$ Race $\square$ Nationality $\square$ Age $\square$ Political $\square$ Religion $\square$ Disability $\square$ Anti-masker \\ $\square$ Anti-vaxxer $\bigcirc$ No identity\end{tabular}                         \\
\midrule
\begin{tabular}[c]{@{}l@{}}Q1B: Are the any individuals mentioned explicitly by their name?\\ $\bigcirc$ Yes $\bigcirc$ No\end{tabular}                                                                                                                                                                                \\
\midrule 
\begin{tabular}[c]{@{}l@{}}Q2: Are you sure that the options you selected in Q1A are indeed identities and NOT just\\ entities such as ``her/she/they'', CCP, Chinese army, USA government, etc.?\\ $\bigcirc$ Yes, I'm sure. $\bigcirc$ No, I need to revise my Q1A's answer.\end{tabular}                            \\
\midrule 
\begin{tabular}[c]{@{}l@{}}Q3: Are there any derogatory, humiliating, insulting, or disparaging words or phrases mentio-\\ned in the text?\\ $\bigcirc$ Yes $\bigcirc$ No\end{tabular}                                                                                                                                   \\
\midrule 
\begin{tabular}[c]{@{}l@{}}Q4A: If Q3's answer is ``Yes'', are those words or phrases directed towards or targeting your\\ selected identities? Select ``No'' if Q3's answer is ``No''.\\ $\bigcirc$ Yes $\bigcirc$ No\end{tabular}                                                                                \\
\midrule 
\begin{tabular}[c]{@{}l@{}}Q4B: If Q3's answer is ``Yes'', are those words or phrases directed towards or targeting the\\ mentioned individuals? Select ``No'' if Q3's answer is ``No''.\\ $\bigcirc$ Yes $\bigcirc$ No\end{tabular}                                                                                 \\
\midrule 
\begin{tabular}[c]{@{}l@{}}Q5: If Q4A or Q4B's answer is ``Yes'', are you sure that the identities or individuals selected\\ are really being attacked? Or is it possible that the text is mere criticism of an entity?\\ $\bigcirc$ Yes, I'm sure. $\bigcirc$ No, I need to revise my Q4's answer.\end{tabular}\\
\bottomrule
\end{tabular}}
\caption{Code book developed by two experts.}
\label{tab:codebook}
\vspace{-5mm}
\end{table}

\section{}
\label{app:samples}

Table~\ref{tab:moresamples} depicts additional samples from our dataset, including the corresponding HateCoT answers and their respective ground truth labels.

\begin{table*}[h]
\centering
\resizebox{!}{1.2\columnwidth}{%
\begin{tabular}{llc}
\toprule
\multicolumn{1}{c}{\textbf{Input Text}} & \multicolumn{1}{c}{\textbf{HateCoT Answers}} & \multicolumn{1}{c}{\textbf{\begin{tabular}[c]{@{}c@{}}Ground\\ Truth\end{tabular}}} \\
\midrule
\begin{tabular}[c]{@{}l@{}}when i print that russians are debils it gets always corrected to devils actually not\\ big difference\end{tabular}                                                                                                                                                            & \begin{tabular}[c]{@{}l@{}}A1: The identity mentioned in the text is `nationality' (Russians).\\ A1B: No, there are no individuals mentioned explicitly by their name.\\ A2: Yes, there are derogatory words used in the text (``debils'' and ``devils'').\\ A3A: Yes, the derogatory words are directed towards the Russians.\\ A3B: Not applicable, no individual is mentioned by name.\\ A4A: Yes, the comment seemingly attempts to incite hate against Russians.\\ A4B: Not applicable, no individual is mentioned by name.\\ A5A: Yes, considering the words used and the nationality they are directed \\ towards, the comment can be concluded as identity hate speech.\\ A5B: Not applicable, no individual is mentioned by name\end{tabular}                                                                                                                                                                                                                                                                              & hate                                                                                \\
\midrule
\begin{tabular}[c]{@{}l@{}}yeh sala china communist leftist maoist pig terrorist country and their dirty pig \\ terrorist chinese people r dangerous to the whole world and humanity these \\ chinese pig terrorist always spreading chinese virus and again a new chinese virus\end{tabular}             & \begin{tabular}[c]{@{}l@{}}A1: The identities mentioned in the text are `nationality' and `political'.\\ A1B: No individuals are mentioned explicitly by their name.\\ A2: Yes, there are several insulting or disparaging words specifically used in \\ the text such as ``sala'', ``communist'', ``leftist'', ``maoist'', ``pig'', ``terrorist'', ``dirty'', \\ ``dangerous'', ``chinese virus''.\\ A3A: Yes, these words or phrases are directed towards or targeting the selected \\ identities which are `nationality' and `political'.\\ A3B: No, there are no words or phrases directed towards or targeting individuals.\\ A4A: Yes, these terms incite hate against the selected identities which are Chinese\\  people and they are referred to as communists, leftists, and maoists.\\ A4B: Not applicable as there are no individuals targeted.\\ A5A: Since the terms used in the comment incite hate against the nationality and\\ political identities, it can be concluded as `Identity Hate'.\\ A5B: Not applicablet.\end{tabular} & hate                                                                                \\
\midrule

\begin{tabular}[c]{@{}l@{}}every antimasker in public is a drunk driver killing Canadians treat them with the \\ same respect you treat any drunk driving through a school zone they don't care if\\ they destroy your life by killing your loved ones\end{tabular}                                        & \begin{tabular}[c]{@{}l@{}}A1: The identity mentioned in the text is `(anti-)masker'.\\ A1B: No individuals are mentioned explicitly by their names.\\ A2: Yes, 'drunk driver killing Canadians' and `they don't care if they destroy your\\ life by killing your loved ones' are derogatory and insulting phrases.\\ A3A: Yes, those words or phrases are directed towards the `(anti-)masker' identity.\\ A3B: No, words or phrases are not directed towards named individuals.\\ A4A: Yes, these terms promote negative attitudes and insights hate against the\\ `(anti-)masker' identity\\ A4B: Not applicable, as no named individuals are insulted or disparaged.\\ A5A: Considering the above answers, the comment is identity hate speech.\\ A5B: Not applicable, as no named individuals are insulted or disparaged.\end{tabular}                                                                                                                                                                                      & hate                                                                                \\
\midrule

\begin{tabular}[c]{@{}l@{}}i like a good conspiracy theory jfk moon landings roswell even boris trump and \\ putin but aren t even real conspiracies because they can be proven basically it s a\\ load of idiots who are just plainly dumb\end{tabular}                                                  & \begin{tabular}[c]{@{}l@{}}A1: The identities mentioned in the text are `political'.\\ A1B: Yes, individuals mentioned explicitly by their names are JFK, Boris, Trump\\ and Putin.\\ A2: Yes, the phrases ``a load of idiots'' and ``just plainly dumb'' can be considered\\ derogatory or insulting.\\ A3A: No, these phrases are not explicitly directed towards the political identity.\\ A3B: Yes, these phrases could be seen as directed towards the individuals,\\ especially since it is in the context of conspiracy theories involving those individuals.\\ A4A: Not Applicable\\ A4B: No, the comment does not necessarily incite hate against these individuals.\\ A5A: Not Applicable\\ A5B: No, the comment cannot be concluded as individual hate speech since it does \\ not purposely incite hate against the individuals mentioned.\end{tabular} & Non-hate \\

\midrule
the only people whit fealings are chinese people & \begin{tabular}[c]{@{}l@{}}A1: The identity mentioned here is `nationality' (Chinese).\\ A1B: No, there aren't any individuals mentioned explicitly by their name.\\ A2: No, there aren't any derogatory, humiliating, insulting, or disparaging words\\ or phrases used in the text.\\ A3A: Not applicable as A2's answer is `No'.\\ A3B: Not applicable as A2's answer is `No'.\\ A4A: Not applicable as A3A's answer is `No'.\\ A4B: Not applicable as A3B's answer is `No'.\\ A5A: Not applicable as A4A's answer is `No', so the final conclusion is `Non-hate'.\\ A5B: Not applicable as A4B's answer is `No', so the final conclusion is `Non-hate'.\end{tabular} & Non-hate \\
\midrule

\begin{tabular}[c]{@{}l@{}}hey i know how to engage teens in our synagogue have them be the purell \\ dispensers since theyre likely to not get covid19 and also nobody wants them \\ around anyway its genius  mandated by an okboomer if i had to guess or like\\ try something else maybe\end{tabular} & \begin{tabular}[c]{@{}l@{}}A1: The identities mentioned in the text are `age' and `religion'.\\ A1B: No, there are no individuals mentioned explicitly by name.\\ A2: Yes, the phrase ``nobody wants them around anyway'' can be seen as \\ derogatory or insulting.\\ A3A: Yes, the derogatory phrases are directed towards the identities `age' (teens).\\ A3B: No, the derogatory words or phrases are not directed toward individuals.\\ A4A: The terms are not exactly inciting hate but rather belittling the `age' group,\\ which is not necessarily the same as hate speech.\\ A4B: Not applicable, there are no derogatory terms towards individuals.\\ A5A: Given that the comment is more belittling and does not necessarily incite\\ hate,  the final conclusion would be `Non-hate'.\\ A5B: Not applicable, as there are no derogatory terms towards individuals.\end{tabular} & Non-hate \\

\bottomrule

\end{tabular}
}
\caption{Additional samples from our dataset.}
\label{tab:moresamples}
\end{table*}


\clearpage
\newpage 

\section{Meta-Review}

The following meta-review was prepared by the program committee for the 2024
IEEE Symposium on Security and Privacy (S\&P) as part of the review process as
detailed in the call for papers.

\subsection{Summary}
Online hate speech is hard to detect due to its rapid changes. This problem only gets amplified by the fact that deploying new detection models can take a long time as relevant data is data collection and a new model is trained. To bridge the gap, the authors proposed the HateGuard framework that depends on LLMs/Chain-of-Though (CoT) reasoning and updates in policy, rather than model, as a rapid response to hate speech in news events. With this framework, the spread of online hate speech can be reduced while a more robust model is developed.

\subsection{Scientific Contributions}
\begin{itemize}
\item Provides a New Data Set For Public 
\item Use Creates a New Tool to Enable Future Science 
\item Provides a Valuable Step Forward in an Established Field
\end{itemize}

\subsection{Reasons for Acceptance}
The paper provides a valuable step forward to a known problem. Hate speech detection is challenging due to its evolving content and temporal dependencies. 
Unlike prior approaches, which largely depend on training models on new data, the CoT-based approach reduces the time-to-deployment to help mitigate the spread of hate speech. Though the approach does not solve the hate speech detection problem, it does help minimize the spread of hate speech while more robust solutions are being developed.




\end{appendices}

\end{document}